\documentclass{article}

\PassOptionsToPackage{numbers}{natbib}

\usepackage[preprint]{neurips_2023}

\usepackage[utf8]{inputenc} %
\usepackage[T1]{fontenc}    %
\usepackage{hyperref}       %
\usepackage{url}            %
\usepackage{booktabs}       %
\usepackage{amsfonts}       %
\usepackage{nicefrac}       %
\usepackage{microtype}      %
\usepackage{xcolor}         %
\usepackage{graphicx}
\usepackage{subcaption}
\usepackage{colortbl}
\usepackage{todonotes}
\usepackage[normalem]{ulem}
\usepackage{bm}
\usepackage{tikz}
\usepackage{pgfplots}
\usepackage{amsmath}
\usepackage{pifont}
\newcommand{\cmark}{\ding{51}}%
\newcommand{\xmark}{\ding{55}}
\usepackage{multirow}

\usepackage{makecell}
\usepackage{wrapfig}
\usepackage{resizegather}
\usepackage{enumitem}
\usepackage{amssymb}

\title{T1: Scaling Diffusion Probabilistic Fields to High-Resolution on Unified Visual Modalities}

\author{%
  Kangfu Mei\\
  Johns Hopkins University \\
  \texttt{kmei1@jhu.edu}\\
  \And
  Mo Zhou\\
  Johns Hopkins University \\
  \texttt{mzhou32@jhu.edu}\\
  \And
  Vishal M. Patel\\
  Johns Hopkins University \\
  \texttt{vpatel36@jhu.edu}\\ 
}

\begin{document}

\maketitle

\begin{abstract}
Diffusion Probabilistic Field (DPF)~\cite{zhuang_diffusion_2023} models the distribution of continuous functions defined over metric spaces.
While DPF shows great potential for unifying data generation of various modalities including images, videos, and 3D geometry, it does not scale to a higher data resolution.
This can be attributed to the ``scaling property'', where it is difficult for the model to capture local structures through uniform sampling.
To this end, we propose a new model comprising of a view-wise sampling algorithm to focus on local structure learning, and incorporating additional guidance, \emph{e.g.}, text description, to complement the global geometry.
The model can be scaled to generate high-resolution data while unifying multiple modalities.
Experimental results on data generation in various modalities demonstrate the effectiveness of our model, as well as its potential as a foundation framework for scalable modality-unified visual content generation.

\end{abstract}

\section{Introduction}

Generative tasks~\cite{rombach_high-resolution_2022, ramesh2022hierarchical} are overwhelmed by diffusion probabilistic models that hold state-of-the-art results on most modalities like audio, images, videos, and 3D geometry.
Take image generation as an example, a typical diffusion model~\cite{ho2020denoising} consists of a forward process for sequentially corrupting an image into standard noise, a backward process for sequentially denoising a noisy image into a clear image, and a score network that learns to denoise the noisy image.

The forward and backward processes are agnostic to different data modalities; however, the architectures of the existing score networks are not.
The existing score networks are highly customized towards a single type of modality, which is challenging to adapt to a different modality.
For example, a recently proposed multi-frame video generation network~\cite{ho_video_2022, ho_imagen_2022} adapting single-frame image generation networks involve significant designs and efforts in modifying the score networks.
Therefore, it is important to develop a unified model that works across various modalities without modality-specific customization, in order to extend the success of diffusion models across a wide range of scientific and engineering disciplines, like medical imaging (\emph{e.g.}, MRI, CT scans) and remote sensing (\emph{e.g.}, LiDAR).

\begin{figure}[t]
    \centering
\includegraphics[width=\textwidth]{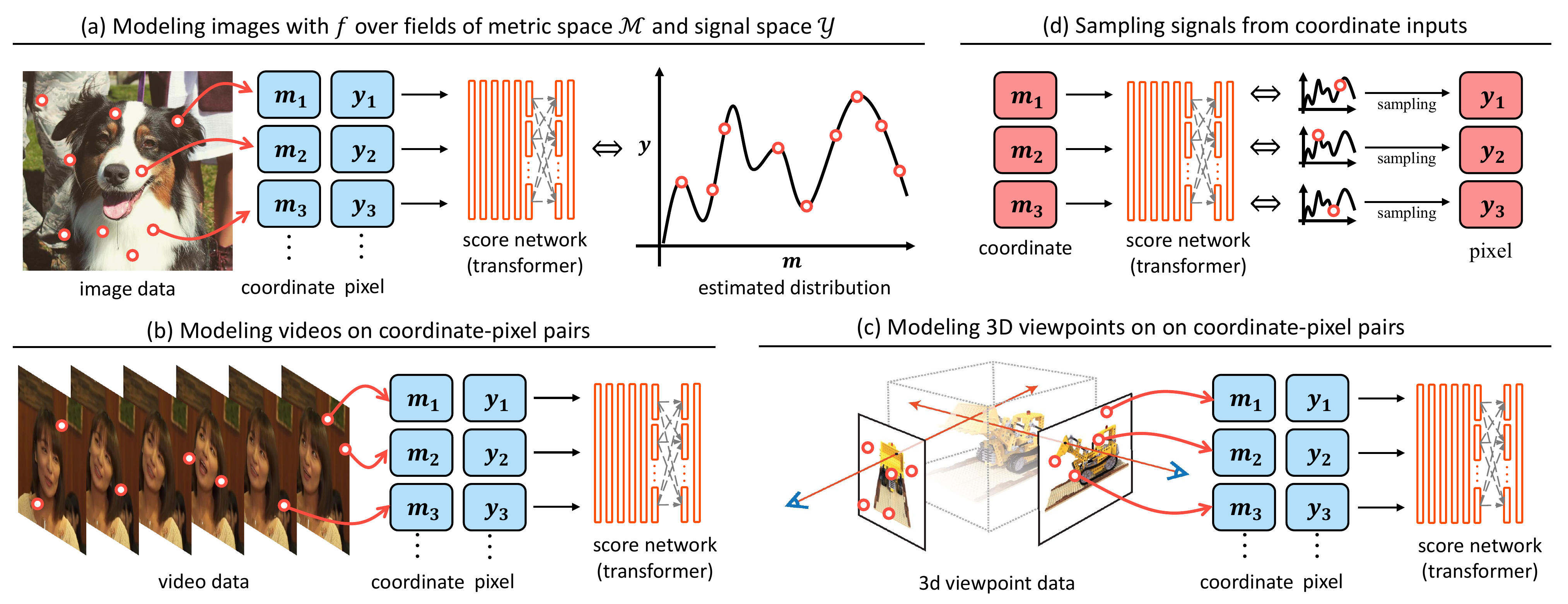}
\vspace{-1.5em}
    \caption{Illustration of the field models' capability of modeling visual content distributions. %
    The %
    underlying data distribution is simplified into the 1-D space for demonstration.
    The score network learns the %
    distribution through the attention among %
    coordinate-signal pairs, which is %
    modality-agnostic.}
    \label{fig:ndf}
    \vspace{-1em}
\end{figure}

Field model~\cite{sitzmann2020implicit, tancik2020fourier, dupont2022data, zhuang_diffusion_2023} is a promising unified score network architecture for different modalities.
It learns the distribution over the functional view of data.
Specifically, the field $f$ maps the observation from the \emph{metric} space $\mathcal{M}$  (\emph{e.g.}, coordinate or camera pose) into the \emph{signal} space $\mathcal{Y}$  (\emph{e.g.}, RGB pixel) as $f:\mathcal{M}\mapsto \mathcal{Y}$.
For instance, an image is represented as $f:\mathbb{R}^2 \mapsto \mathbb{R}^3$ that maps the spatial coordinates (\emph{i.e.}, height and width) into RGB values at the corresponding location (See Fig.~\ref{fig:ndf} (a)),
while a video is represented as $f:\mathbb{R}^3 \mapsto \mathbb{R}^3$ that maps the spatial and temporal coordinates (\emph{i.e.}, frame, height, and width) into RGB values (See Fig.~\ref{fig:ndf} (b)).
Recently, diffusion models are leveraged to characterize the field distributions over the functional view of data~\cite{zhuang_diffusion_2023} for field generation.
Given a set of coordinate-signal pairs $\{(\bm{m}_i,\bm{y}_i)\}$,
the field $f$ is regarded as the score network for the backward process,
which turns a noisy signal into a clear signal $\bm{y}_i$ in a sequential process with $\bm{m}_i$ being fixed all the time, as shown in Fig.~\ref{fig:ndf} (d).
The visual content is then composed of the clear signal generated on a grid in the metric space.

Nevertheless, diffusion-based field models for generation still lag behind the modality-specific approaches~\cite{dhariwal2021diffusion, ho_video_2022, he_latent_2022} for learning from dynamic data in high resolution~\cite{bain2021frozen, yu2023celebv}.
For example, a 240p video lasting 5 seconds is comprised of up to 10 million coordinate-signal pairs. 
Due to the memory bottleneck in existing GPU-accelerated computing systems, recent field models~\cite{zhuang_diffusion_2023} are limited to observe merely a small portion of these pairs (\emph{e.g.}, $1\%$) that are uniformly sampled during training.
This limitation significantly hampers the field models in approximating distributions from such sparse observations~\cite{quinonero2005unifying}.
Consequently, diffusion-based field models often struggle to capture the fine-grained local structure of the data, leading to, \emph{e.g.}, unsatisfactory blurry results.

While it is possible to change the pair sampling algorithm to sample densely from local areas instead of uniformly,
the global geometry is weakened.
To alleviate this issue, it is desirable to introduce some complementary guidance on the global geometry in addition to local sampling.

Multiple attempts~\cite{Gordon2020Convolutional, dutordoir2022neural, zhuang_diffusion_2023} have been presented to introduce additional global priors during modeling.
Recent diffusion models~\cite{rombach_high-resolution_2022, ramesh2022hierarchical} demonstrate that text descriptions can act as strong inductive biases for learning data distributions, by introducing global geometry priors of the data, thereby helping one to scale the models on complex datasets.
However, fully exploiting correlation between the text and the partially represented field remains uncharted in the literature.

In this paper, we aim to address the aforementioned issues, and scale the field models for generating high-resolution, dynamic data.
We propose a new diffusion-based field model, called \textbf{T1}.
In contrast to previous methods, T1 preserves both the local structure and the global geometry of the fields during learning by employing a new view-wise sampling algorithm in the coordinate space, and incorporates additional inductive biases from the text descriptions.
By combining these advancements with our simplified network architecture, we demonstrate that T1's modeling capability  surpasses previous methods, achieving improved generated results under the same memory constraints.
We empirically validate its superiority against previous domain-agnostic methods across three different tasks, including image generation, text-to-video generation, and 3D viewpoint generation.
Various experiments show that T1 achieves compelling performance even when compared to the state-of-the-art domain-specific methods, underlining its potential as a scalable and unified visual content generation model across various modalities.
Notably, T1 is capable of generating high-resolution video under affordable computing resources, while the existing field models can not.

Our contributions are summarized as follows:
\begin{itemize}[leftmargin=2em,noitemsep]
    \item We reveal the scaling property of diffusion-based field models, which prevents them from scaling to high-resolution, dynamic data despite their capability of unifying various visual modalities.
    \item We propose T1, a new diffusion-based field model with a sampling algorithm that maintains the view-wise consistency, and enables the incorporation of additional inductive biases.
\end{itemize}

\section{Background}

Conceptually, the diffusion-based field models sample from field distributions by reversing a gradual noising process.
As shown in Fig.~\ref{fig:ndf}, in contrast to the data formulation of the conventional diffusion models~\cite{ho2020denoising} applied to the complete data like a whole image, diffusion-based field models apply the noising process to the sparse observation of the field, which is a kind of parametrized functional representation of data consisting of coordinate-signal pairs, \emph{i.e.}, $f:\mathcal{M}\mapsto \mathcal{Y}$.
Specifically, the sampling process begins with a coordinate-signal pair $(\mathbf{m}_i, \mathbf{y}_{(i, T)})$, where the coordinate comes from a field and the signal is a standard noise, and less-noisy signals $\mathbf{y}_{(i, T-1)}, \mathbf{y}_{(i, T-2)},\dots,$ are progressively generated until reaching the final clear signal $\mathbf{y}_{(i, 0)}$, with $\mathbf{m}_i$ being constant.

Diffusion Probabilistic Field (DPF)~\cite{zhuang_diffusion_2023} is one of the recent representative diffusion-based field models.
It parameterizes the denoising process with a transformer-based network $\epsilon_\theta(\cdot)$, which takes noisy coordinate-signal pairs as input and predicts the noise component $\epsilon$ of $\mathbf{y}_{(i, t)}$.
The less-noisy signal $\mathbf{y}_{(i, t-1)}$ is then sampled from the noise component $\epsilon$ using a denoising process~\cite{ho2020denoising}.
For training, they use a simplified loss proposed by Ho et al.~\cite{ho2020denoising} instead of the variational lower bound for modeling the distributions in VAE~\cite{kingma2013auto}.
Specifically, it is a simple mean-squared error between the true noise and the predicted noise, \emph{i.e.}, $\|\epsilon_\theta(\mathbf{m}_i, \mathbf{y}_{(i, t)}, t) - \epsilon\|$.
This approach is found better in practice and is equivalent to the denoising score matching model~\cite{song2020improved}, which belongs to another family of denoising models and is referred to as the denoising diffusion model.

In practice, when handling low-resolution data consisting of $N$ coordinate-signal pairs with DPF, the scoring network $\epsilon_\theta(\cdot)$ takes all pairs $\{(\mathbf{m}_i, \mathbf{y}_{(i, T)})\}$ as input at once. 
For high-resolution data with a large number of coordinate-signal pairs that greatly exceed the modern GPU capacity, Zhuang et al.~\cite{zhuang_diffusion_2023} uniformly sample a subset of pairs from the data as input.
They subsequently condition the diffusion model on the other non-overlapping subset, referred to as \emph{context pairs}.
Specifically, the sampled pairs interact with the query pairs through cross-attention blocks.
Zhuang et al.~\cite{zhuang_diffusion_2023} show that the ratio between the context pairs and the sampling pairs is strongly related to the quality of the generated fields, and the quality decreases as the context pair ratio decreases.
In this paper, we show that the context pairs fail to present high-resolution, dynamic data. Thus, we propose a new sampling algorithm along with the conditioning mechanism for scaling the diffusion-based field models.

\section{Method}
In order to scale diffusion-based field models for high-resolution, dynamic data generation, we build upon the recent DPF model~\cite{zhuang_diffusion_2023} and address its limitations in preserving the local structure of fields, as it can hardly be captured when the uniformly sampled coordinate-signal pairs are too sparse.
Specially, our method not only can preserve the local structure, but also enables introducing additional inductive biases (\emph{i.e., text descriptions)} for capturing the global geometry.

\begin{figure}
    \centering
    \includegraphics[width=\textwidth]{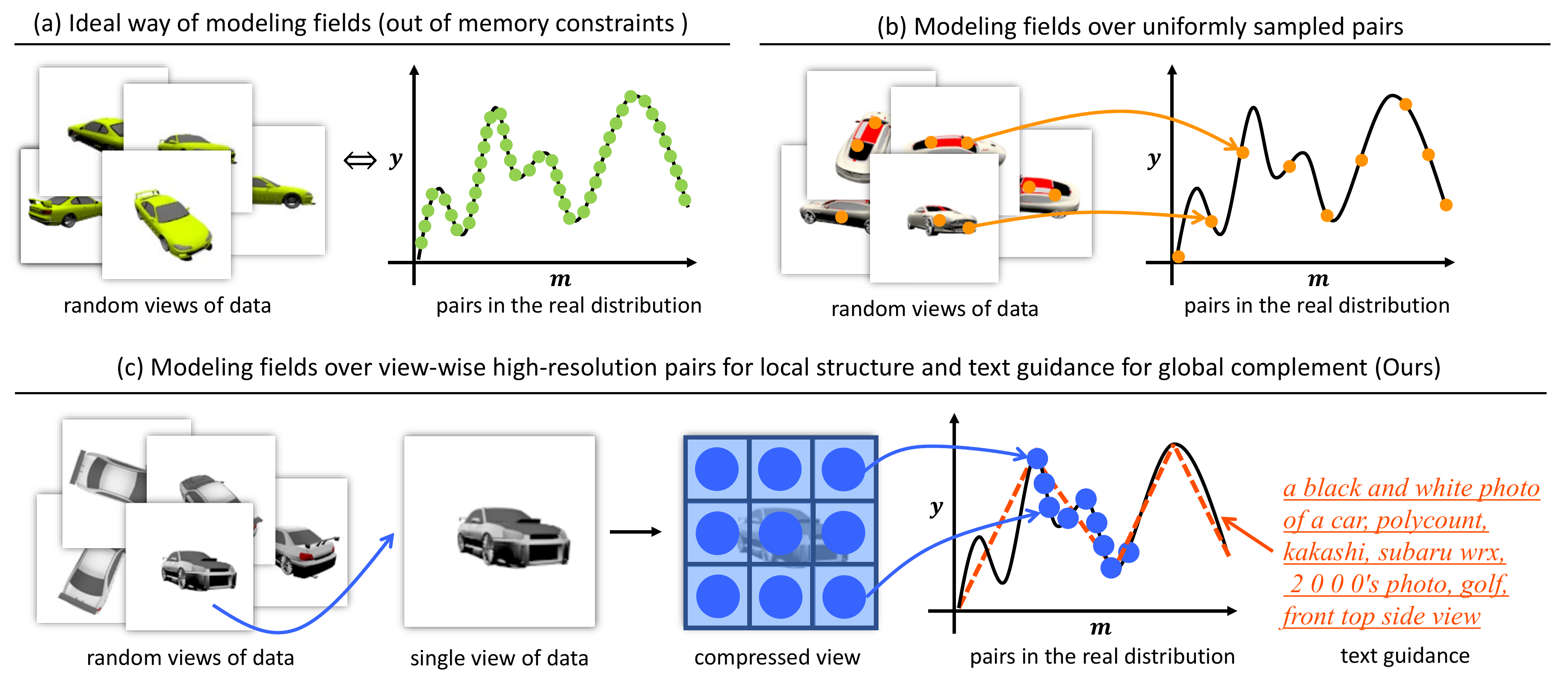}
    \caption{Sampling strategies on high-dimensional data.
    (a) Ideally, all pairs within a field (green points) should be used for training, but is impractical given the memory limitations.
    (b) Previous methods uniformly sample a sparse set of pairs (orange points) to represent the field.
    (c) Compared to uniform sampling, our local sampling extracts high-resolution pairs (blue points), better covering the local structure. The text guidance (red line) as an approximation complements the global geometry.}
    
    \label{fig:overview}
\end{figure}

\subsection{View-wise Sampling Algorithm}

In order to preserve the local structure of fields, we propose a new view-wise sampling algorithm that samples local coordinate-signal pairs for better representing the local structure of fields.
For instance, the algorithm samples the coordinate-signal pairs belonging to a single or several ($n \geqslant 1$; $n$ denotes the number of views) views for video data,
where a view corresponds to a single frame. It samples pairs belonging to a single or several rendered images for 3D viewpoints, where a view corresponds to an image rendered at a specific camera pose.
A view of an image is the image itself.

This approach restricts the number of interactions among pairs to be modeled and reduces the learning difficulty on high-resolution, dynamic data.
Nevertheless, even a single high-resolution view , \emph{e.g.}, in merely $128{\times}128$ resolution) can still consist of 10K pairs, which in practice will very easily reach the memory bottleneck if we leverage a large portion of them at one time, and hence hinder scaling the model for generating high-resolution dynamic data.

To address this issue, our method begins by increasing the signal resolution of coordinate-signal pairs and reducing memory usage in the score network.
Specifically, we replace the signal space with a compressed latent space, and employ a more efficient network architecture that only contains decoders.
This improvement in efficiency allows the modeling of interactions among pairs representing higher-resolution data while keeping the memory usage constrained.
Based on this, one can then model the interactions of pairs within a single or several views of high-resolution data.
The overall diagram of the proposed sampling algorithm can be found in Fig.~\ref{fig:overview}.

\paragraph{Signal Resolution.} 
We construct the coordinate-signal pairs in a compressed latent space, \emph{i.e.}, each signal is represented by a transformer token, where the signal resolution for each token is increased from $1\times 1\times 3$ to $16\times 16\times 3$ compared to the baseline, while maintaining the memory cost of each pair.
In particular, for each view of the data in a $H \times W \times 3$ resolution, we first extract its latent representation using a pre-trained autoencoder~\cite{rombach_high-resolution_2022}, with the latent map size being $H/8 \times W/8 \times 4$.
This approach improves the field representation efficiency by perceptually compressing the resolution.
We then employ a convolutional layer with $2\times 2$ kernel size in the score network for further compressing the latent, resulting in a compressed feature map in $H/16\times H/16 \times 4$ resolution.
This step further improves the computation efficiency of the scoring network by four times, which is particularly useful for transformers that have quadratic complexity.

In this way, each coordinate-signal pair contains a coordinate, and its corresponding $1\times 1$ feature point (corresponds to a $16\times 16$ signal)
from the compressed feature map (with positional embedding added).
For each token, we use their corresponding feature map location for the position embedding.
By combining these, in comparison to DPF which supports a maximum $64\times 64$ view resolution, our method can handle views with a maximum resolution of $1024 \times 1024$ while maintaining very close memory consumption during learning without compromising the quality of the generated signal.

\paragraph{Score Network.}
We further find that once a token encapsulates enough information to partially represent the fidelity of the field, the context pairs~\cite{zhuang_diffusion_2023} are no longer necessary for model efficiency.
Therefore, using high-resolution tokens enables us to get rid of the encoder-decoder architecture~\cite{jaegle2021perceiver} and thus to utilize a more parameters-efficient decoder-only architecture.
We adopt DiT~\cite{peebles2022scalable} as the score network, which is the first decoder-only pure-transformer model that takes noisy tokens and positional embedding as input and generates the less-noisy token.

\paragraph{View-wise Sampling Algorithm.}
Based on the high-resolution signal and decoder-only network architecture, our method represents field distributions by using view-consistent coordinate-signal pairs, \emph{i.e.}, collections of pairs that belong to a single or several ($n\geqslant 1$) views of the data, such as one or several frames in a video, and one or several viewpoints of a 3D geometry.
In particular, take the spatial and temporal coordinates of a video in $H\times W$ resolution lasting for $T$ frames as an example, for all coordinates $\{\mathbf{m}_1, \mathbf{m}_2, \dots, \mathbf{m}_i,\dots, \mathbf{m}_{H\times W \times T}\}$, we randomly sample a consecutive sequence of length $H\times W$ that correspond to a single frame, \emph{i.e.}, $\{\mathbf{m}_1, \mathbf{m}_2, \dots, \mathbf{m}_i,\dots, \mathbf{m}_{H\times W}\}$.
For data consisting of a large amount of views (\emph{e.g.} $T >> 16$), we randomly sample $n$ views (sequences of length $H\times W$), resulting in an $H\times W \times n$ sequence set.
Accordingly, different from the transformers in previous works~\cite{zhuang_diffusion_2023} that model interaction among all pairs across all views, ours only models the interaction among pairs that belongs to the same view, which reduces the complexity of field model by limiting the number of interactions to be learned.

\subsection{Text Conditioning}
To complement our effort in preserving local structures that may weaken global geometry learning, since the network only models the interaction of coordinate-signal pairs in the same view, we propose to supplement the learning with a coarse global approximation of the field, avoiding issues in cross-view consistency like worse spatial-temporal consistency between frames in video generation.

\begin{wrapfigure}{ri}{0.22\textwidth}
\vspace{-1\baselineskip}
\begin{center}
\includegraphics[width=0.22\textwidth]{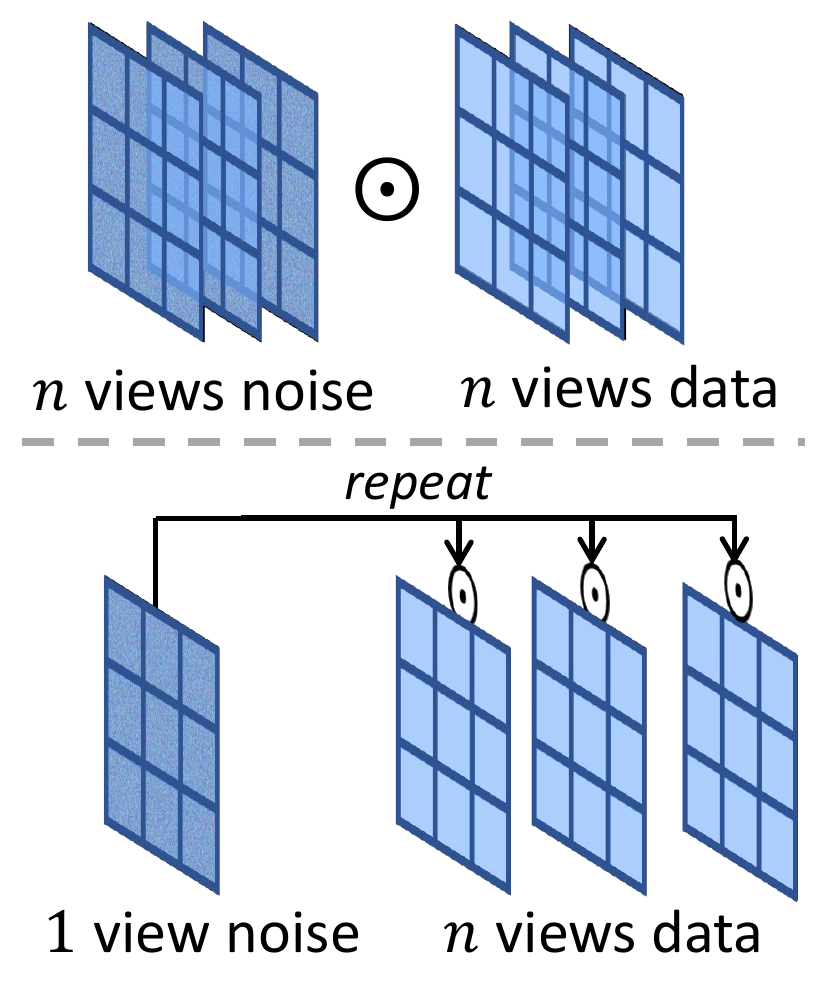}
\end{center}
\vspace{-1\baselineskip}
\caption{Overview of the previous noisy data construction (top) and ours (bottom).}
\vspace{-2\baselineskip}
\label{fig:noise}
\end{wrapfigure}

In particular, we propose to condition diffusion models on text descriptions related to the fields.
Compared with the other possible modalities, text can better represent data in compact but highly expressive features~\cite{devlin2018bert,brown2020language,raffel2020exploring}, and serve as a low-rank approximation of data~\cite{radford2021learning}.
By conditioning diffusion models on text descriptions, we show our method can capture the global geometry of data from texts.
It works like inductive biases of each pairs and allow us to model cross-view interactions of pairs without explicit cross-attention used in previous methods~\cite{zhuang_diffusion_2023}.

\paragraph{Cross-view Noise Consistency.}
We propose to model the interactions among pairs across different views, which indeed represent the dependency between views as the global geometry.
In particular, we perform the forward diffusion process that constructs cross-view noisy pairs by using the same noise component across views, as illustrated in Fig.~\ref{fig:noise}.
The reparameterization trick~\cite{kingma2013auto} (for the forward process) is then applied to a set of sampled pairs $\mathbf{Q}$ of a field, where the pairs make up multiple views, as shown below:
\begin{equation}
\begin{split}
  \mathbf{Q} &= \Big\{
  \underbrace{ \{ (\mathbf{m}_i, \mathbf{y}_{(i,t)}) | i=1,2,\ldots,H{\cdot}W \}}_{\text{pairs from the $n$-th view}}
  ~~\big|~~ n = 1, 2, \ldots, N
  \Big\} \\
  & = \Big\{\{ (\mathbf{m}_{(i, n)}, \mathbf{y}_{(i, n,t)} = \sqrt{\bar \alpha} \mathbf{y}_{(i, n, 0)} + \sqrt{1-\bar \alpha_t} \epsilon_i) | i{=}1,2,\ldots,H{\cdot}W \}
  ~~\big|~~ n{=}1, 2, \ldots, N
  \Big\}.\\
\end{split}
\end{equation}

In contrast to the previous works that use different noise components for all views of a field, ours results in a modified learning objective, \emph{i.e.}, to coherently predict the same noise component from different distorted noisy views.
In this way, the whole field is regarded as a whole where each view is correlated with the others.
This enforces the model to learn to generate coherent views of a field.

\paragraph{Cross-view Condition Consistency.}
In order to model the dependency variation between views belonging to the same field, \emph{i.e.}, the global geometry of the field, we condition the diffusion model on the text embeddings of the field description or equivalent embeddings (\emph{i.e.}, the language embedding of a single view in the CLIP latent space~\cite{radford2021learning}).
Our approach leverages the adaptive layer normalization layers in GANs~\cite{brock2018large, karras2019style}, and adapts them by modeling the statistics from the text embeddings of shape $Z\times D$.
For pairs that make up a single view, we condition on their represented tokens $Z\times D$, ($Z$ tokens of size $D$), by modulating them with the scale and shift parameters regressed from the text embeddings.
For pairs $(T\times Z) \times D$ that make up multiple views, we condition on the view-level pairs by modulating feature in $Z\times D$ for each of the $T$ views with the same scale and shift parameters.
Specifically, each transformer blocks of our score network learns to predict statistic features $\beta_c$ and $\gamma_c$ from the text embeddings per channel. These statistic features then modulate the transformer features $F_c$ as:
$    \mathrm{adLNorm}(F_c | \beta_c, \gamma_c) = \mathrm{Norm}(F_c) \cdot \beta_c + \beta_c $.

\section{Experimental Results}

We demonstrate the effectiveness of our method on multiple modalities, including 2D image data on a spatial metric space $\mathbb{R}^2$, 3D video data on a spatial-temporal metric space $\mathbb{R}^3$, and 3D viewpoint data on a camera pose and intrinsic parameter metric space $\mathbb{R}^6$, while the score network implementation remains identical across different modalities, except for the embedding size.
The concrete network implementation details including architecture and hyper-parameters can be found in the appendix.

\begin{figure}[t]
    \centering
    \begin{subfigure}{.49\textwidth}
    \includegraphics[width=\textwidth]{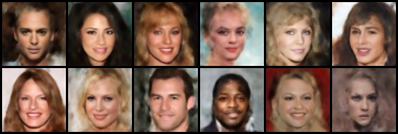}
    \caption{GEM~\cite{du2021learning}}
    \end{subfigure}
    \begin{subfigure}{.49\textwidth}
    \includegraphics[width=\textwidth]{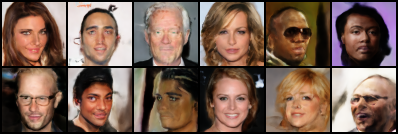}
    \caption{GASP~\cite{dupont_data_2022}}
    \end{subfigure}
    \hfill
    
    \begin{subfigure}{.49\textwidth}
    \includegraphics[width=\textwidth]{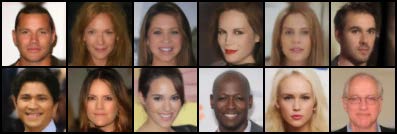}
    \caption{DPF~\cite{zhuang_diffusion_2023}}
    \end{subfigure}
    \begin{subfigure}{.49\textwidth}
    \includegraphics[width=\textwidth]{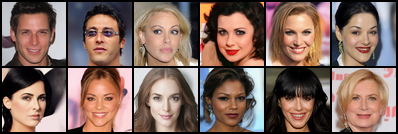}
    \caption{\textbf{T1} (Ours)}
    \end{subfigure}
    \hfill
    \caption{Qualitative comparisons of domain-agnostic methods and ours on CelebA-HQ. Our results show better visual quality with more details than the others, while being domain-agnostic as well.}
    \vspace{-1\baselineskip}
    \label{fig:2dimage}
\end{figure}

\begin{table}[t]
    \setlength\tabcolsep{4pt}
    \begin{center}
    \resizebox{\linewidth}{!}{
    \begin{tabular}[t]{rcccccccccc}
    \toprule
    \multirow{2.5}{*}{\bf Model} & \multicolumn{3}{c}{\bf{CelebA-HQ} 64$\times$64} & \multicolumn{3}{c}{\bf{CelebV-Text} 256$\times$256$\times$128} & \multicolumn{4}{c}{\bf{ShapeNet-Cars} 128$\times$128$\times$251}  \\
    \cmidrule(r){2-4} \cmidrule(r){5-7} \cmidrule(r){8-11}
     & FID ($\downarrow$) & Pre. ($\uparrow$) & Rec. ($\uparrow$) & FVD ($\downarrow$) & FID ($\downarrow$) & CLIPSIM ($\uparrow$) & FID ($\downarrow$) & LPIPS ($\downarrow$) & PSNR ($\uparrow$) & SSIM ($\uparrow$) \\
    \midrule
    Functa~\cite{dupont_data_2022} & 40.40 & 0.58 & 0.40 & \xmark & \xmark & \xmark & 80.3 & N/A & N/A & N/A \\
    GEM~\cite{du2021learning} & 30.42 & 0.64 & 0.50 & \xmark & \xmark & \xmark & \xmark & \xmark & \xmark & \xmark  \\
    GASP~\cite{dupont2021generative} & 13.50 & 0.84 & 0.31 & \xmark & \xmark & \xmark & \xmark & \xmark & \xmark & \xmark  \\
    DPF~\cite{zhuang_diffusion_2023} & 13.21 & \textbf{0.87} & 0.35 & \xmark & \xmark & \xmark & \xmark & \xmark & \xmark & \xmark \\
    \midrule
    TFGAN~\cite{balaji2019conditional} & \xmark & \xmark & \xmark & 571.34 & 784.93 & 0.154 & \xmark & \xmark & \xmark & \xmark  \\
    MMVID~\cite{han_show_2022} & \xmark & \xmark & \xmark & 109.25 & 82.55 & 0.174 & \xmark & \xmark & \xmark & \xmark \\
    MMVID-interp~\cite{han_show_2022} & \xmark & \xmark & \xmark & 80.81 & 70.88 & 0.176 & \xmark & \xmark & \xmark & \xmark \\
    VDM~\cite{ho_video_2022} & \xmark & \xmark & \xmark & 81.44 & 90.28 & 0.162 & \xmark & \xmark & \xmark & \xmark \\
    CogVideo~\cite{hong2022cogvideo} & \xmark & \xmark & \xmark & 99.28 & 54.05 & 0.186 & \xmark & \xmark & \xmark & \xmark \\
    \midrule
    EG3D-PTI~\cite{chan2022efficient} & \xmark & \xmark & \xmark & \xmark & \xmark & \xmark & 20.82 & 0.146 & 19.0 & 0.85 \\
    ViewFormer~\cite{kulhanek2022viewformer} & \xmark & \xmark & \xmark & \xmark & \xmark & \xmark & 27.23 & 0.150 & 19.0 & 0.83 \\
    pixelNeRF~\cite{yu2021pixelnerf} & \xmark & \xmark & \xmark & \xmark & \xmark & \xmark & 65.83 & 0.146 & 23.2 & 0.90 \\
    \midrule
    \textbf{T1 (Ours)} & \textbf{5.55} & 0.77 & \textbf{0.51} & \textbf{42.03} & \textbf{24.33} & \textbf{0.220} & 24.36 & \textbf{0.118} & \textbf{23.9} & \textbf{0.90}  \\
    \bottomrule
    \end{tabular}
    }
    \end{center}
    \caption{Sample quality comparison with state-of-the-art models for each task. ``\xmark'' denotes the method cannot be adopted to the modality due to the method design or impractical computation cost.}
    \vspace{-1\baselineskip}
    \label{tab:sota}
\end{table}

\begin{figure}[htbp]
    \centering
    \begin{subfigure}[b]{1\textwidth}
    \centering
    \includegraphics[width=.13\textwidth]{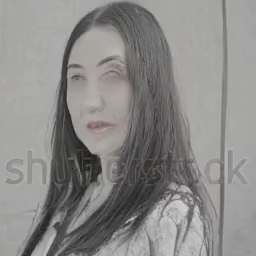}
    \includegraphics[width=.13\textwidth]{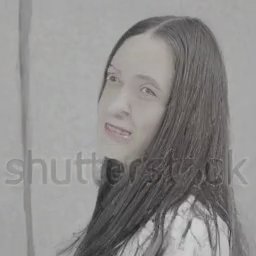}
    \includegraphics[width=.13\textwidth]{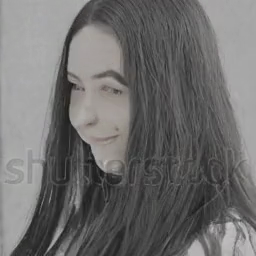}
    \includegraphics[width=.13\textwidth]{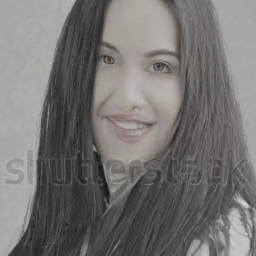}
    \includegraphics[width=.13\textwidth]{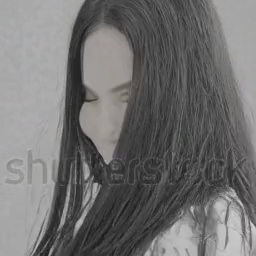}
    \includegraphics[width=.13\textwidth]{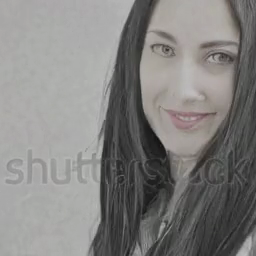}
    \includegraphics[width=.13\textwidth]{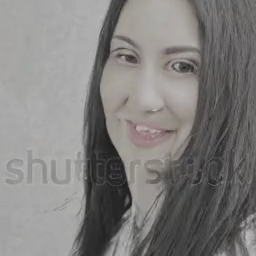}
    \hfill
    \caption{VDM~\cite{ho_video_2022}}
    \end{subfigure}
    
    \begin{subfigure}[b]{1\textwidth}
    \centering
    \includegraphics[width=.13\textwidth]{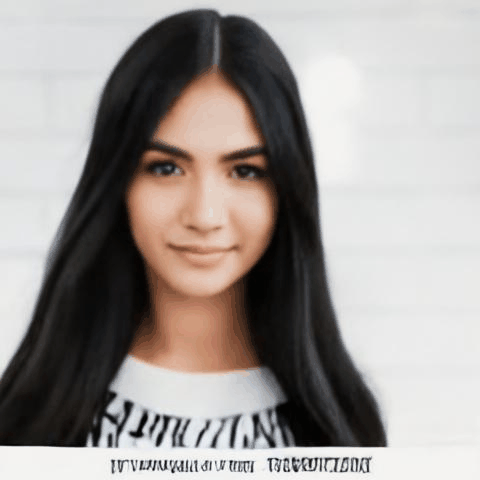}
    \includegraphics[width=.13\textwidth]{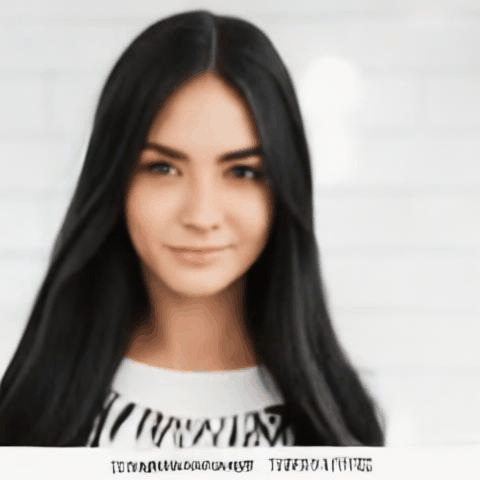}
    \includegraphics[width=.13\textwidth]{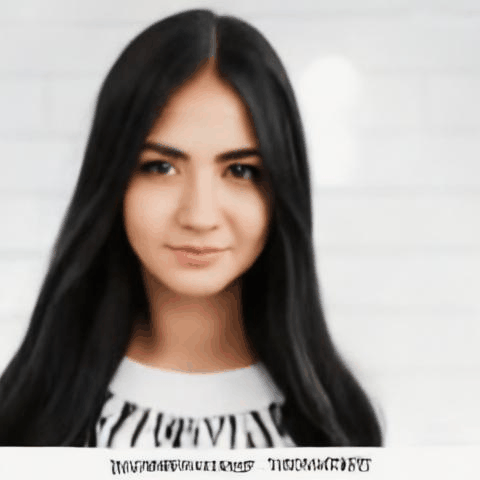}
    \includegraphics[width=.13\textwidth]{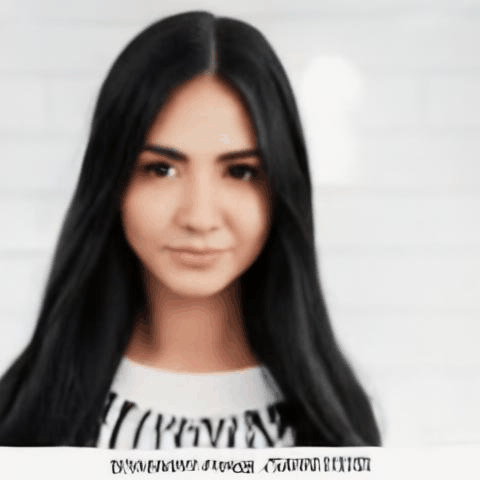}
    \includegraphics[width=.13\textwidth]{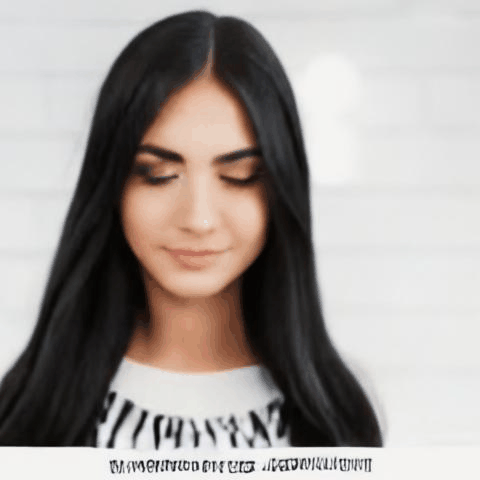}
    \includegraphics[width=.13\textwidth]{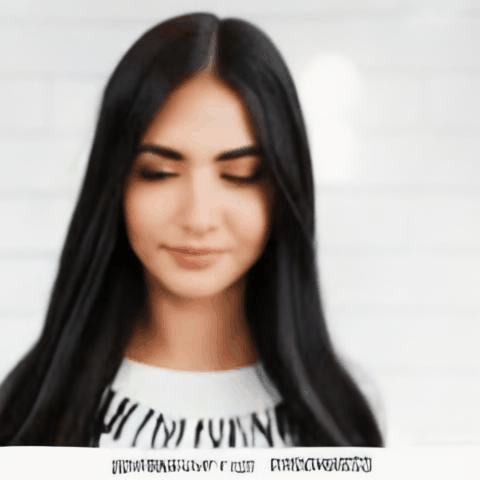}
    \includegraphics[width=.13\textwidth]{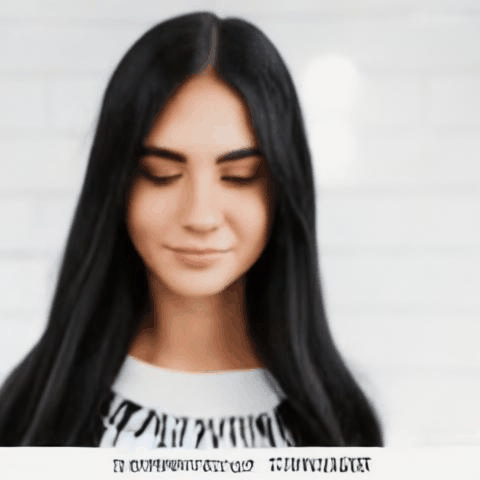}
    \hfill
    \caption{CogVideo~\cite{hong2022cogvideo}}
    \end{subfigure}

    \begin{subfigure}[b]{1\textwidth}
    \centering
    \centering
    \includegraphics[width=.13\textwidth]{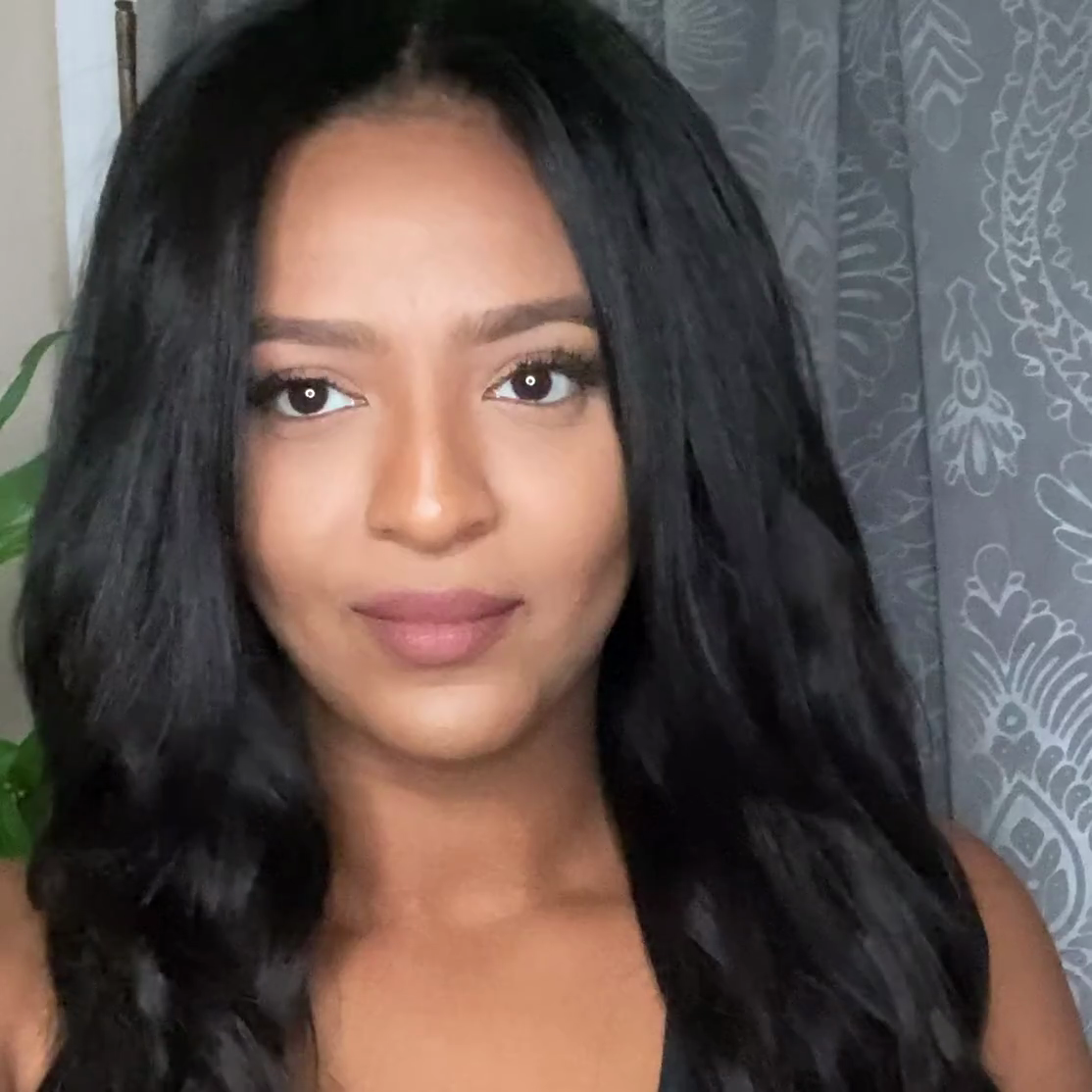}
    \includegraphics[width=.13\textwidth]{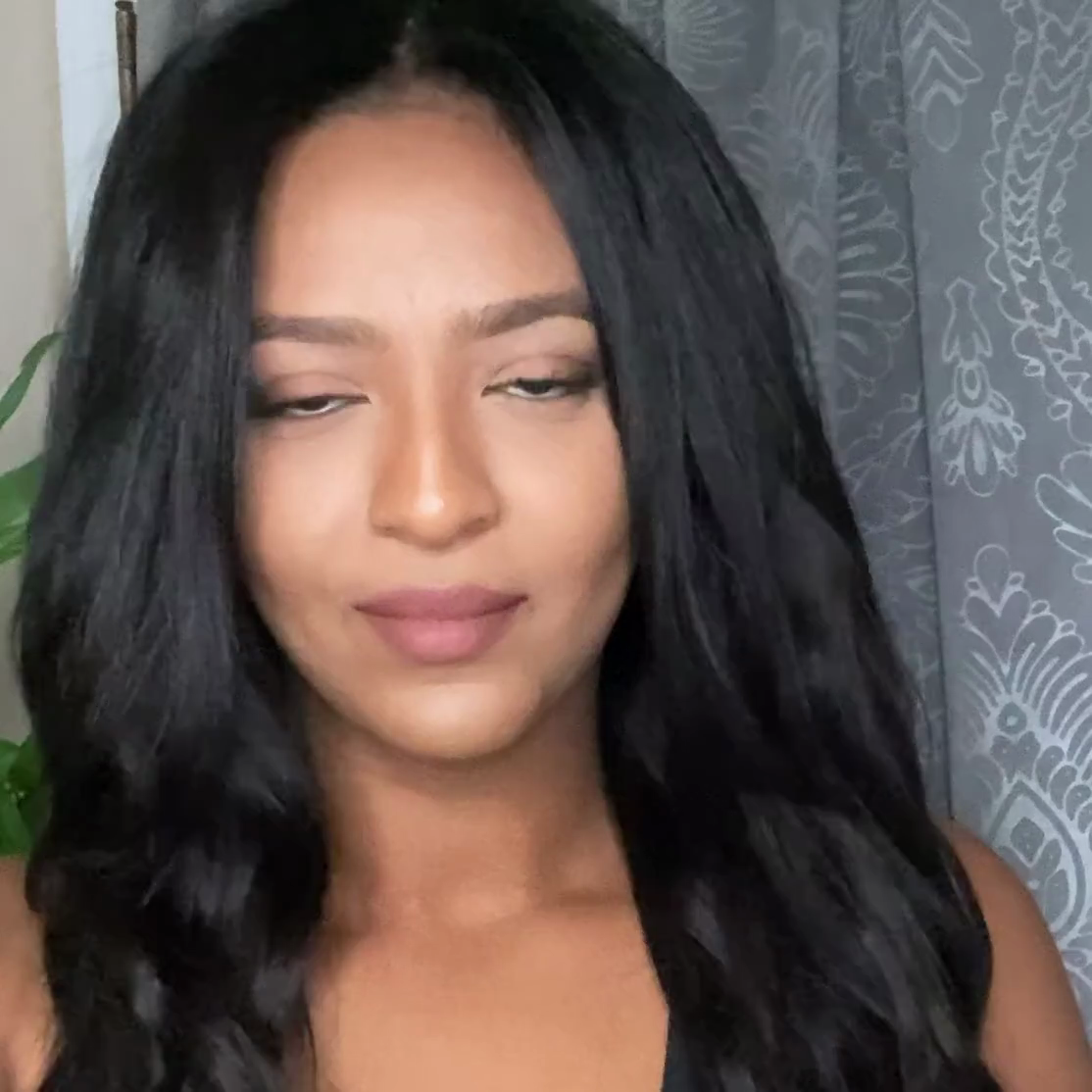}
    \includegraphics[width=.13\textwidth]{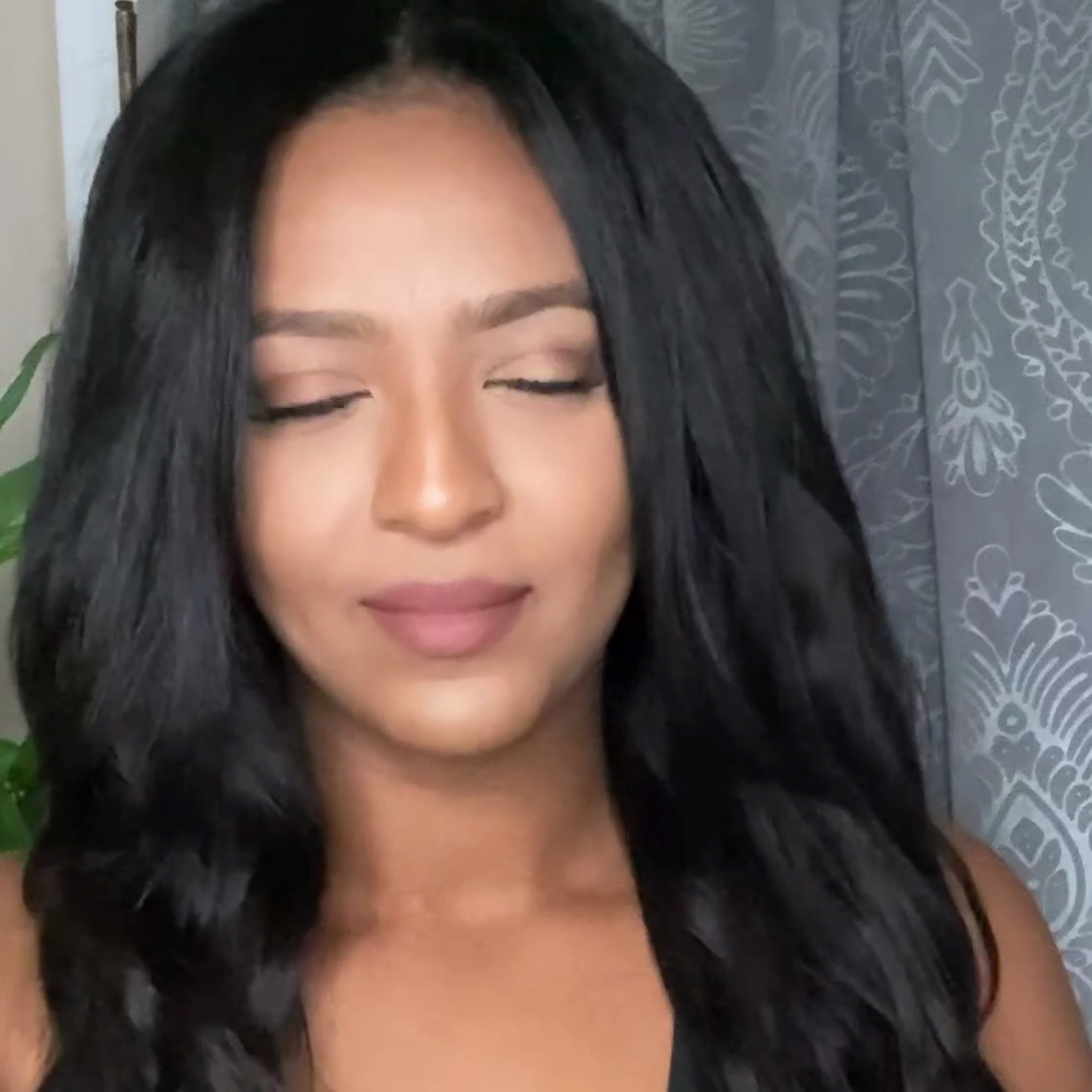}
    \includegraphics[width=.13\textwidth]{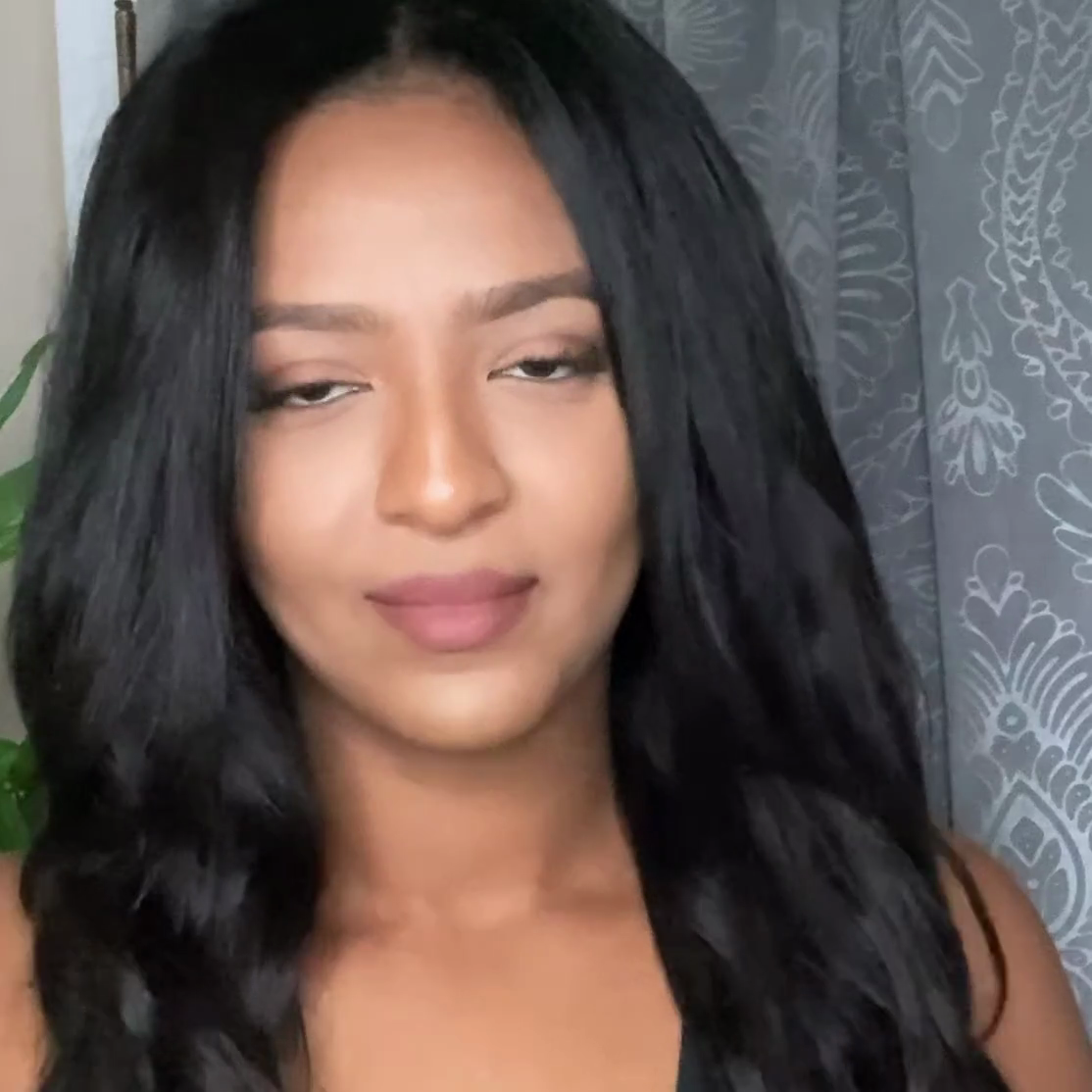}
    \includegraphics[width=.13\textwidth]{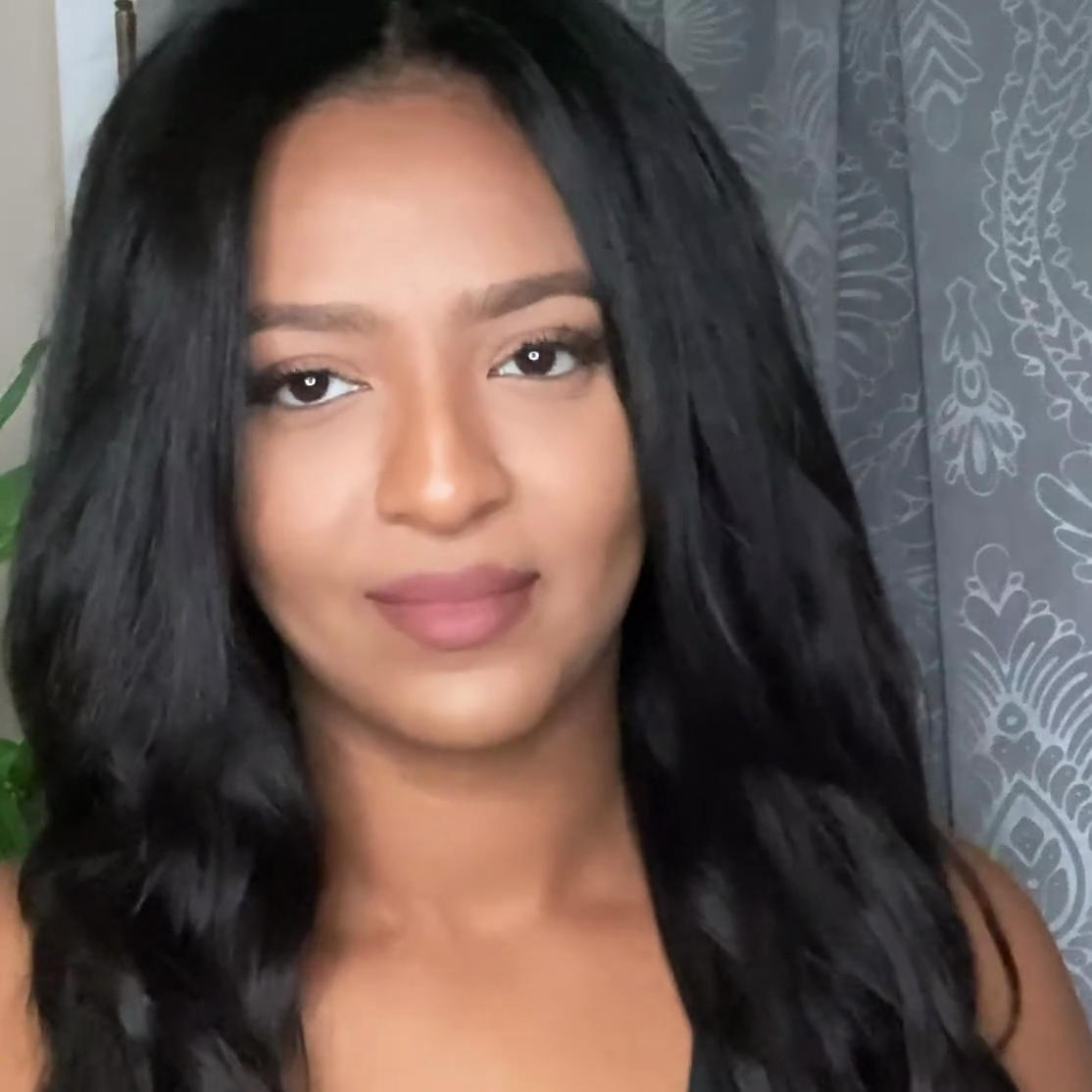}
    \includegraphics[width=.13\textwidth]{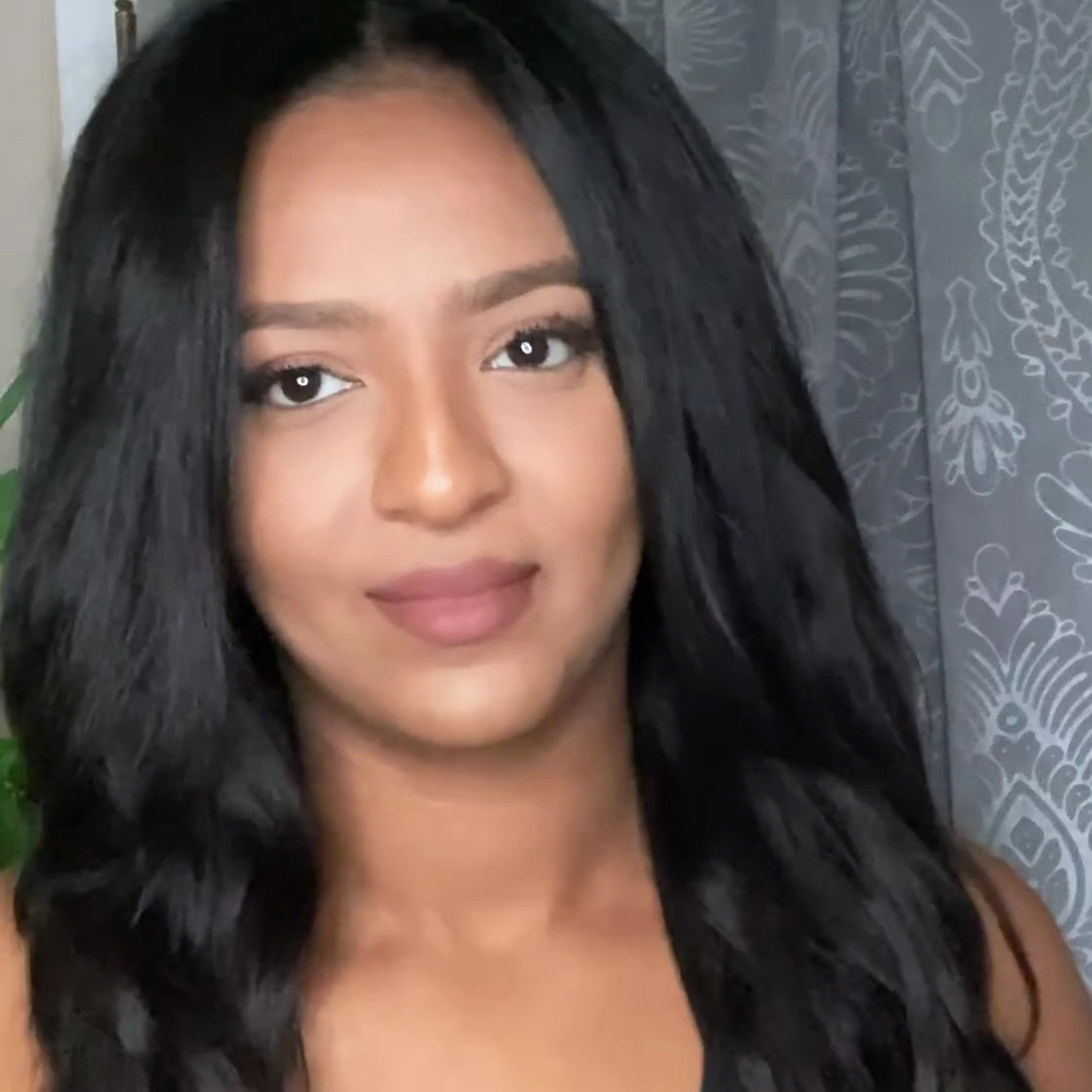}
    \includegraphics[width=.13\textwidth]{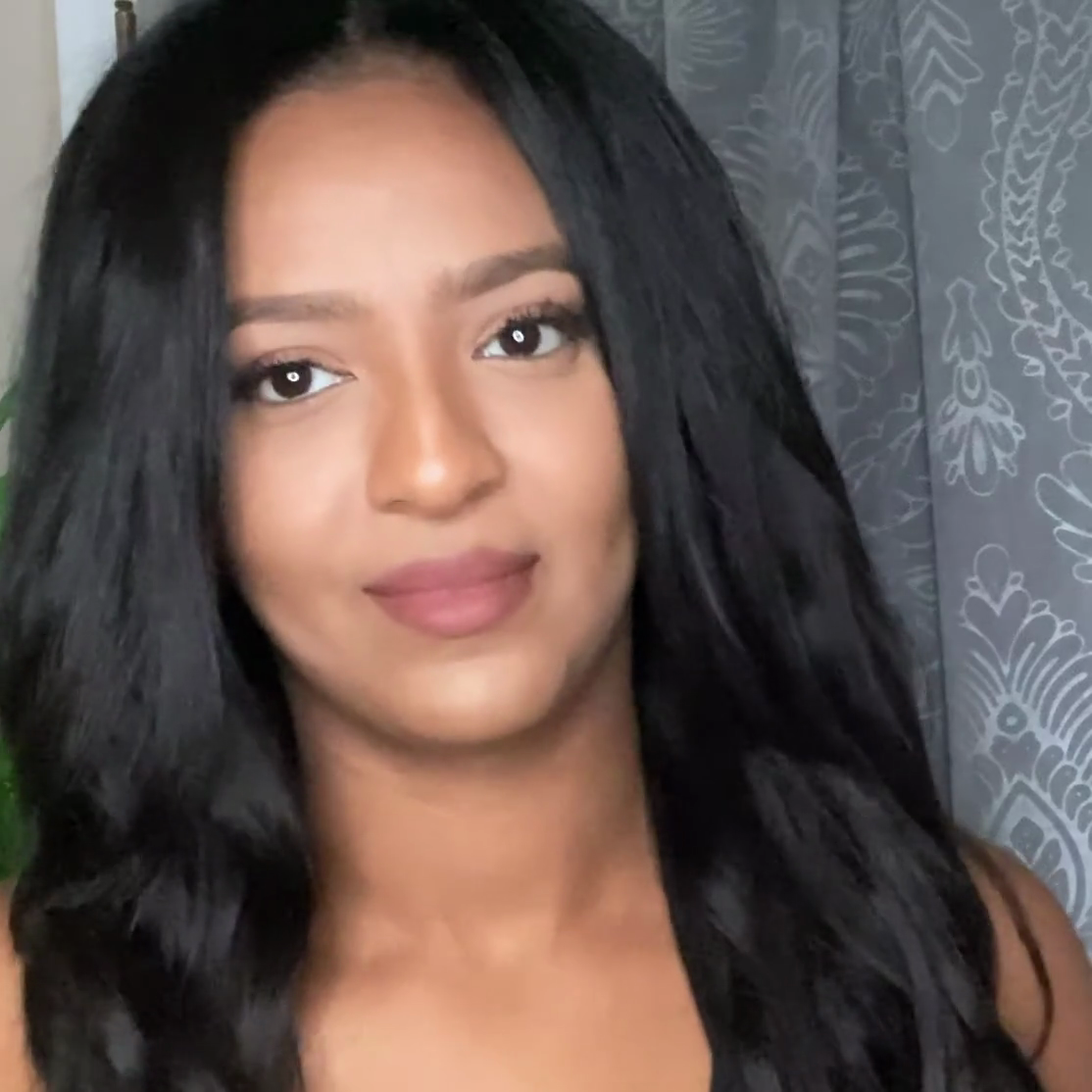}
    \caption{\textbf{T1} (Ours)}
    \end{subfigure}
    \caption{Qualitative comparisons of domain-specific text-to-video models and ours with prompt: ``\emph{This young female has straight hair. She has long black hair. The woman has arched eyebrows and bags under eyes. She smiles all the time.}'' Compared with VDM~\cite{ho_video_2022}, our result is more continuous. Compared with CogVideo~\cite{hong2022cogvideo}, our result have more realistic textures, \emph{e.g.}, of the hair.}
    \label{fig:video}
\end{figure}

\paragraph{Images.}
For image generation, we use the standard benchmark dataset, \emph{i.e.}, CelebA-HQ 64$\times$64~\cite{liu2015deep, karras2017progressive} as a sanity test, in order to compare with other domain-agnostic and domain-specific methods.
For the low-resolution CelebA-HQ dataset, we compare our method with the previous domain-agnostic methods including DPF~\cite{zhuang_diffusion_2023}, GASP~\cite{dupont2021generative}, GEM~\cite{du2021learning}, and Functa~\cite{dupont2022data}.
We report Fréchet Inception Distance (FID)~\cite{heusel2017gans} and Precision/Recall metrics~\cite{kynkaanniemi2019improved} for quantitative comparisons~\cite{sajjadi2018assessing}.

The experimental results can be found in Tab.~\ref{tab:sota}.
Specifically, T1 outperforms all domain-agnostic models in the FID metric and Recall score, while achieving a very competitive Precision score.
The difference in our Precision score stems from the usage of ImageNet pretraining~\cite{peebles_scalable_2022}, which affects the diversity of the generated data as well as its distribution, instead of the generated image quality.
The qualitative comparisons in Fig.~\ref{fig:2dimage} further demonstrate our method's superiority in images.

\paragraph{Videos.}
To show our model's capacity for more complex data, \emph{i.e.}, high-resolution, dynamic video, we conduct experiments on the recent text-to-video
benchmark: CelebV-Text 256$\times$256$\times$128~\cite{yu_celebv-text_2023} (128 frames).
As additional spatial and temporal coherence is enforced compared to images, video generation is relatively underexplored by domain-agnostic methods.
We compare our method with the representative domain-specific methods including TFGAN~\cite{balaji2019conditional}, MMVID~\cite{han2022show}, CogVideo~\cite{hong2022cogvideo} and VDM~\cite{ho_video_2022}.
We report Fréchet Video Distance (FVD)~\cite{unterthiner2018towards}, FID, and CLIPSIM~\cite{wu2021godiva}, \emph{i.e.}, the cosine similarity between the CLIP embeddings~\cite{radford2021learning} of the generated images and the corresponding texts.
Note, the recent text-to-video models (like NUAW~\cite{wu2022nuwa}, Magicvideo~\cite{zhou2022magicvideo}, Make-a-video~\cite{singer2022make}, VLDM~\cite{blattmann2023align}, etc.) are not included in our comparisons. This is solely because all of them neither provide implementation details, nor runnable code and pretrained checkpoints. Furthermore, their approaches are similar to VDM~\cite{ho_video_2022}, which is specifically tailored for video data.

Our method achieves the best performance in both the video quality (FVD) and signal frame quality (FID) in Tab.~\ref{tab:sota}, compared with the recent domain-specific text-to-video models.
Moreover, our model learns more semantics as suggested by the CLIPSIM scores.
The results show that our model, as a domain-\emph{agnostic} method, can achieve a performance on par with domain-\emph{specific} methods in the generation of high-resolution, dynamic data.
The qualitative comparisons in Fig.~\ref{fig:video} further support our model in text-to-video generation compared with the recent state-of-the-art methods.

\begin{figure}[tbp]
    \centering
    \begin{subfigure}{.31\textwidth}
    \includegraphics[width=\textwidth]{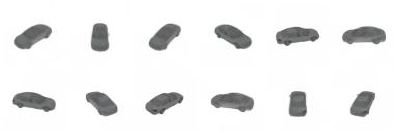}
    \caption{pixelNeRF~\cite{yu2021pixelnerf}}
    \end{subfigure}
    \hfill
    \begin{subfigure}{.31\textwidth}
    \includegraphics[width=\textwidth]{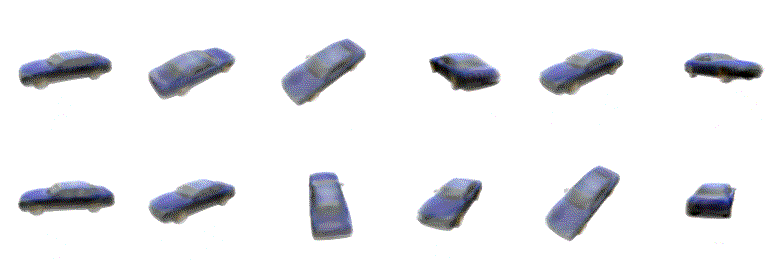}
    \caption{Functa~\cite{dupont2022data}}
    \end{subfigure}    
    \hfill
    \begin{subfigure}{.31\textwidth}
    \includegraphics[width=\textwidth]{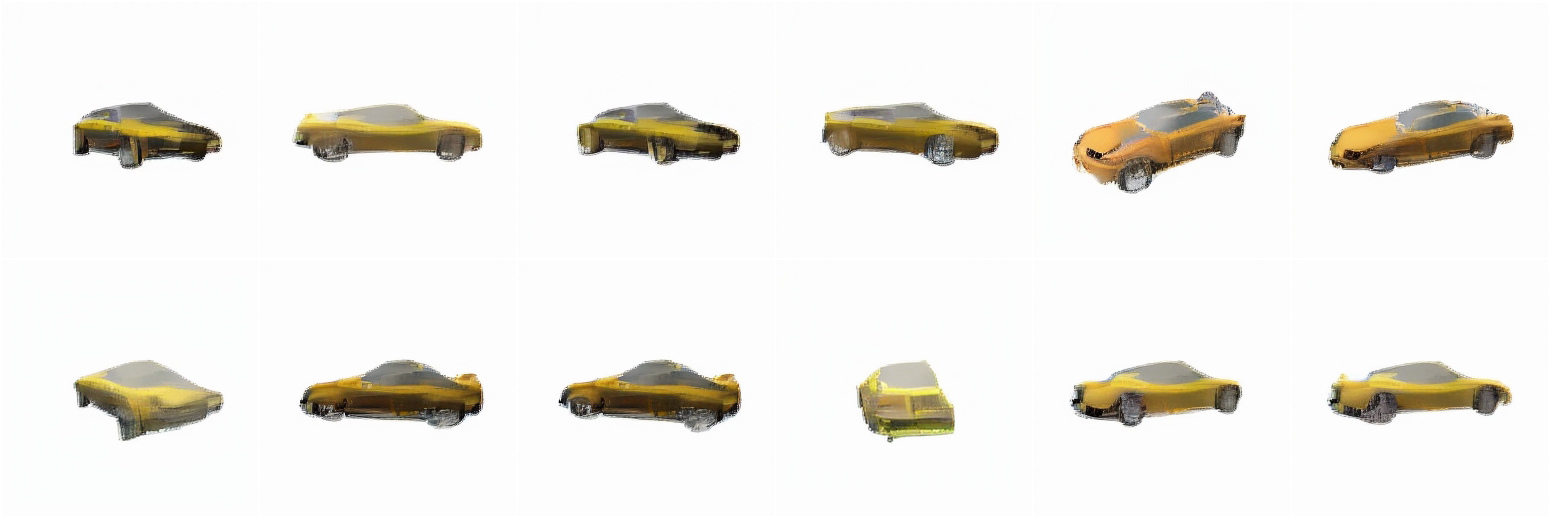}
    \caption{\textbf{T1} (Ours)}
    \end{subfigure}
    \hfill

    \begin{subfigure}{\textwidth}
    \includegraphics[width=\textwidth]{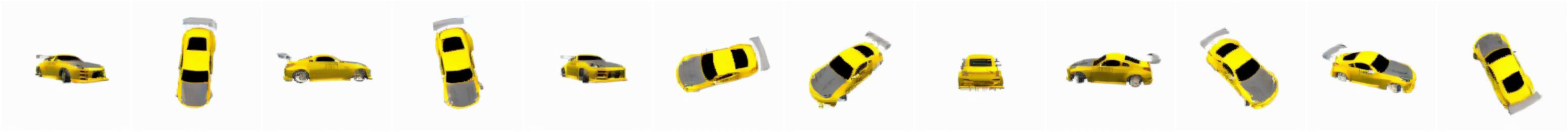}
    \caption{\textbf{T1} (Our high-resolution result)}
    \end{subfigure}
    \hfill
    
    \caption{Qualitative comparisons of domain-specific novel view synthesis models and ours. Our results show competitive quality without explicitly using 3D modeling and allows generating high-resolution results (\emph{e.g.}, 256$\times$256$\times$251 ) by only using low-resolution training data.}
    \vspace{-1\baselineskip}
    \label{fig:3d}
\end{figure}

\paragraph{3D Viewpoints.}
We also evaluate our method on 3D viewpoint generation with the ShapeNet dataset~\cite{chang2015shapenet}.
Specifically, we use the ``car'' class of ShapeNet which involves 3514 different cars.
Each car object has 50 random viewpoints, where each viewpoint is in 128 $\times$ 128 resolution.
Unlike previous domain-agnostic methods~\cite{du2021learning, zhuang_diffusion_2023} that model 3D geometry over voxel grids at $64^3$ resolution, we model over rendered camera views based on their corresponding camera poses and intrinsic parameters, similar to recent domain-specific methods~\cite{sitzmann2019scene, yu2021pixelnerf}.
This approach allows us to extract more view-wise coordinate-signal pairs while voxel grids only have 6 views.
We report our results in comparison with the state-of-the-art view-synthesis algorithms including pixelNeRF~\cite{yu2021pixelnerf}, viewFormer~\cite{kulhanek2022viewformer}, and EG3D-PTI~\cite{chan2022efficient}.
Note that our model indeed performs one-shot novel view synthesis by conditioning on the text embedding of a randomly sampled view.

Our model's performance is even comparable with domain-\emph{specific} novel view synthesize methods, as shown by the result in Tab.~\ref{tab:sota}.
Since our model does not explicitly utilize 3D geometry regularization as NeRF does, the compelling results demonstrate the potential of our method across various complex modalities like 3D geometry. 
The visualizations in Fig.~\ref{fig:3d} also show similar quality as previous works.

\subsection{Ablations and Discussions}
In this section, we demonstrate the effectiveness of each of our proposed components and analyze their contributions to the quality of the final result, as well as the computation cost.
The quantitative results under various settings are shown in Table~\ref{tab:ablation}.
Since the text conditioning effect depends on our sampling algorithm, we will first discuss the effects of text conditions and then local sampling.

\begin{table}[t]
    \begin{center}
    \resizebox{1.0\columnwidth}{!}{
    \begin{tabular}{cc|cc|c|ccccc}
    \toprule
     Text   & \thead{View-wise \\ Noise} & Local Sampling & Resolution & \thead{Training \\ Views} & FVD ($\downarrow$) & FID ($\downarrow$) & CLIPSIM ($\uparrow$) & MACs & Mems   \\
    \midrule
    \xmark  & N/A       & \cmark    & 16.0 & 8 & 608.27 & 34.10 & - & 113.31G & 15.34Gb \\
    \cmark  & \xmark    & \cmark    & 16.0 & 8 & 401.64 & 75.81 & 0.198 & 117.06G & 15.34Gb \\
    \hline
    \cmark  & N/A    & \xmark  & 1.0* & 8 & 900.03 & 119.83 & 0.113 & 7.350T & 60.31Gb \\
    \cmark  & \cmark  & \cmark   & 1.0* & 8 & 115.20 &  40.34 & 0.187 & 7.314T & 22.99Gb  \\
    \hline
    \cmark  & \cmark  & \cmark   & 16.0 & 1 & 320.02 & 21.27 & 0.194 & 117.06G & 15.34Gb  \\
    \cmark  & \cmark  & \cmark   & 16.0 & 4 & 89.83 & 23.69 & 0.194 & 117.06G & 15.34Gb  \\
    \hline
    \cmark  & \cmark    & \cmark    & 16.0 & 8 & 42.03 & 24.33 & 0.220 & 117.06G & 15.34Gb \\
    \bottomrule
    \end{tabular}
    }
    \end{center}
    \caption{Ablation analysis on our proposed method under different settings. `*' denotes that the model is trained on low-resolution 32$\times$32 videos due the setting is not efficient enough and reach the memory constrains. All computation cost MACs and GPU memory usage Mems are estimated in generating a single view regardless of the resolution for a fair comparison.}
    \label{tab:ablation}
    \vskip -0.2in
\end{table}

\paragraph{Effect of text condition.} To verify the effectiveness of the text condition for capturing the global geometry of the data, we use two additional settings.
\textbf{(1)} The performance of our model when the text condition is removed is shown in the first row of Tab.~\ref{tab:ablation}.
The worse FVD means that the text condition play a crucial role in preserving the global geometry, specifically the spatial-temporal coherence in videos.
\textbf{(2)} When the text condition is added, but not the cross-view consistent noise, the results can be found in the second row of Tab.~\ref{tab:ablation}.
The FVD is slightly improved compared to the previous setting, but the FID is weakened due to underfitting against cross-view inconsistent noises.
In contrast to our default setting, these results demonstrate the effectiveness of the view-consistent noise.

\paragraph{Effect of local sampling.}
We investigate the effects of the local sampling under different settings for preserving the local structure of data.
\textbf{(1)} We first compare our local sampling with the baseline uniform sampling strategy~\cite{zhuang_diffusion_2023},
as shown in the $3$rd row and $4$th row of Tab.~\ref{tab:ablation}.
Specifically, due to the memory constrains, we can only conduct experiments on frames in a lower resolution of 32$\times$32 during sampling pairs, which are marked with ``*''.
The FID evaluated on single frames shows the local structure quality, and hence the effectiveness of local sampling.
Furthermore, our local sampling significantly reduces memory usages, from 60.31Gb into 22.99Gb, at a $0.036$T less cost of MACs.
\textbf{(2)} To verify the effectiveness of the extended signal resolution, we can compare the $4$th row (resolution $1{\times}1$) and the last row (default setting; resolution $16{\times}16$).
In contrast, our default setting outperforms the low-resolution setting without significant computation and memory consumption.

\begin{figure}[t]
    \centering
    \begin{subfigure}[b]{1\textwidth}
    \centering
    \includegraphics[width=.13\textwidth]{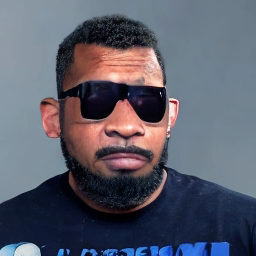}
    \includegraphics[width=.13\textwidth]{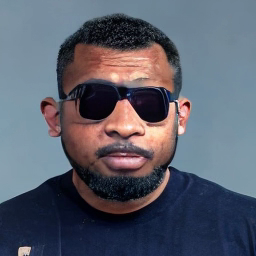}
    \includegraphics[width=.13\textwidth]{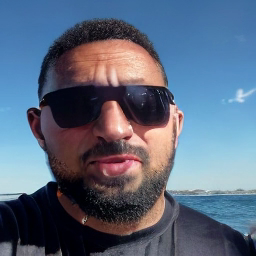}
    \includegraphics[width=.13\textwidth]{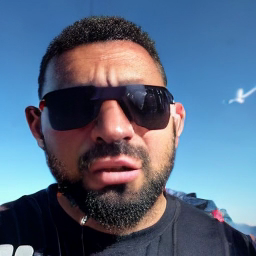}
    \includegraphics[width=.13\textwidth]{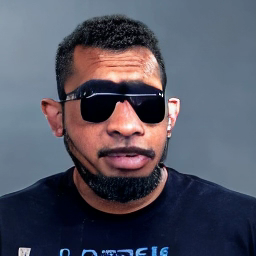}
    \includegraphics[width=.13\textwidth]{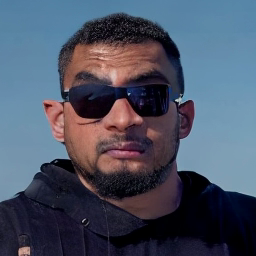}
    \includegraphics[width=.13\textwidth]{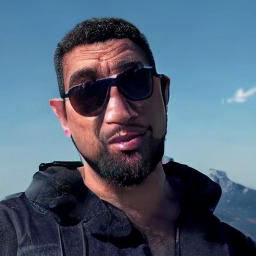}
    \hfill
    \caption{1 views setting with prompt: \emph{``This man is young. He wears eyeglasses. He talks for a moderate time.''}}
    \end{subfigure}

    \begin{subfigure}[b]{1\textwidth}
    \centering
    \includegraphics[width=.13\textwidth]{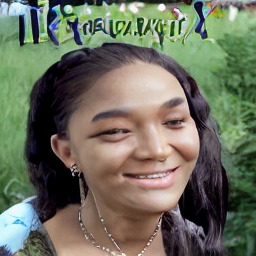}
    \includegraphics[width=.13\textwidth]{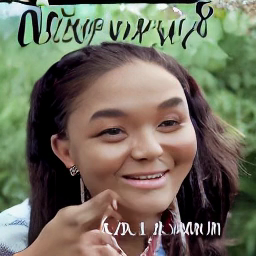}
    \includegraphics[width=.13\textwidth]{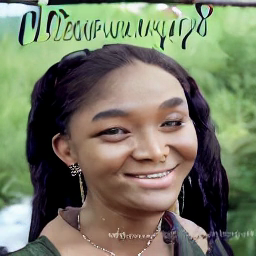}
    \includegraphics[width=.13\textwidth]{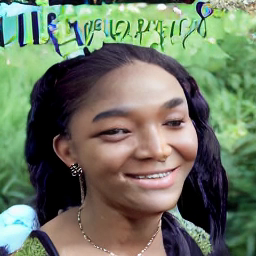}
    \includegraphics[width=.13\textwidth]{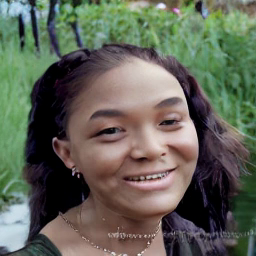}
    \includegraphics[width=.13\textwidth]{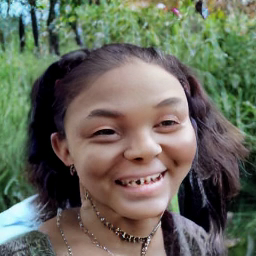}
    \includegraphics[width=.13\textwidth]{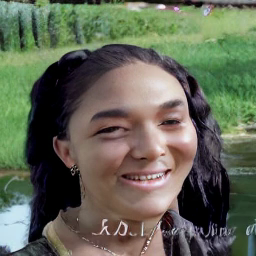}
    \hfill
    \caption{4 views setting with prompt: \emph{``She is young. This woman talks while smiling for a long time.''}}
    \end{subfigure}

    \begin{subfigure}[b]{1\textwidth}
    \centering
    \includegraphics[width=.13\textwidth]{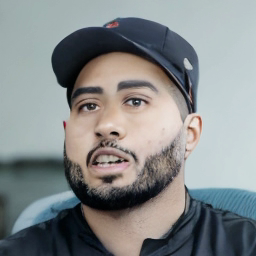}
    \includegraphics[width=.13\textwidth]{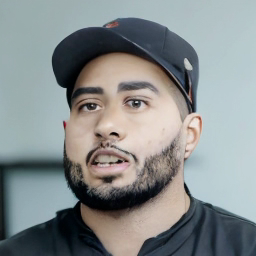}
    \includegraphics[width=.13\textwidth]{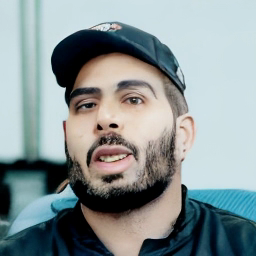}
    \includegraphics[width=.13\textwidth]{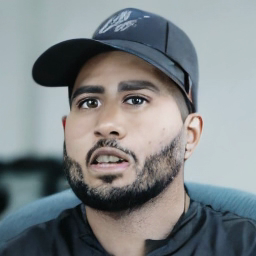}
    \includegraphics[width=.13\textwidth]{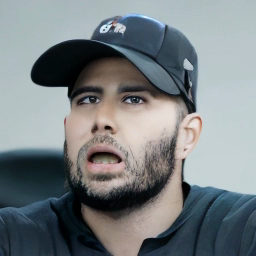}
    \includegraphics[width=.13\textwidth]{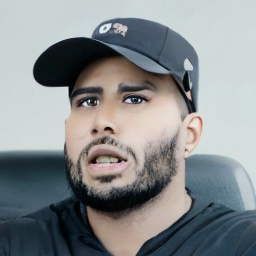}
    \includegraphics[width=.13\textwidth]{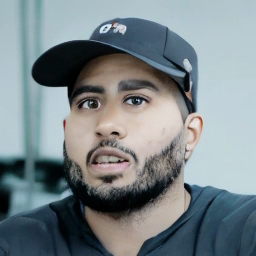}
    \hfill
    \caption{8 views setting with prompt: \emph{``He has beard. The man turns for a short time, then he talks for a short time.''}}
    \end{subfigure}
    
    \caption{Qualitative comparisons of our model with different settings of $n$. The visual results under 1 view setting suffers from periodically identity changes. The visual results under 4 views and 8 views settings gradually improve the identity consistency among frames.}
    \label{fig:abview}
    \vspace{-1\baselineskip}
\end{figure}

\paragraph{Effect of number of views.}
We investigate the model performance change with varying number of views ($n$) for representing fields, as shown in the $5$th and $6$th rows of Tab.~\ref{tab:ablation}.
Compared to the default setting of $n=8$, reducing $n$ to $1$ leads to non-continuous frames and abrupt identity changes, as indicated by the low FVD.
When $n$ is increased to $4$, the continuity between frames is improved, but still worse than the default setting with $n=8$ for the dynamics between frames.
As the $n=8$ setting reaches the memory limit, we set it as the default.
Thus, a larger number of views leads to a higher performance, along with a higher computation cost.
The visualizations are shown in Fig.~\ref{fig:abview}.

\paragraph{Limitations.}
(1) Our method can generate high-resolution data, but the scaling property is merely resolved for the spatial dimensions exclusively.
For instance, for an extremely long video with complex dynamics (\emph{e.g.}, 1 hour; such long videos remain uncharted in the literature), learning short-term variations is still difficult since our local sampling method is still uniform in the temporal perspective. This paper focuses on generating spatially high-resolution data.
(2) Our method only applies to visual modalities interpretable by views. For modalities such as temperature manifold~\cite{hersbach2019era5} where there is no ``views'' of such field, our method does not apply, but Functa~\cite{dupont2022data} does.

\section{Related Work}
In recent years, generative models have shown impressive performance in visual content generation.
The major families are generative adversarial networks~\cite{goodfellow2020generative, mao2017least, karras2019style, brock2018large}, variational autoencoders~\cite{kingma2013auto, vahdat2020nvae}, auto-aggressive networks~\cite{chen2020generative, esser2021taming}, and diffusion models~\cite{ho2020denoising, song2020score}.
Recent diffusion models have obtained significant advancement with stronger network architectures~\cite{dhariwal2021diffusion}, additional text conditions~\cite{ramesh2022hierarchical}, and pretrained latent space~\cite{he_latent_2022}.
Our method built upon these successes and targets at scaling domain-agnostic models for matching these advancement.

Our method models field distributions using explicit coordinate-signal pairs, which is different from the body of work that implicitly models field distributions, including Functa~\cite{dupont2022data} and GEM~\cite{du2021learning}.
These methods employ a two-stage modeling paradigm, which first parameterizes fields and then learns the distributions over the parameterized latent space. 
Compared with the single-stage parameterization used in our method, the two-stage paradigm demands more complex network architecture, as it employs a separate network to formulate a hypernetwork~\cite{ha2016hypernetworks}.
Moreover, the learning efficiency of the two-stage methods hinders scaling the models, as their first stage incurs substantial computational costs to compress fields into latent codes.
In contrast, our method enjoy the benefits of the single-stage modeling and improves its accuracy in preserving local structures and global geometry.

Our method also differs from the recently proposed domain-specific works for high-resolution, dynamic data, which models specific modalities in a dedicated latent space, including Spatial Functa~\cite{bauer2023spatial} and PVDM~\cite{yu2023video}.
These methods typically compress the high-dimensional data into a low-dimensional latent space.
However, the compression is usually specific to a center modality and lacks the flexibility in dealing with different modalities.
For instances, PVDM compresses videos into three latent codes that represent spatial and temporal dimensions separately.
However, such a compressor cannot be adopted into the other similar modalities like 3D scenes.
In contrast, our method owns the unification flexibility by learning on the coordinate-signal pairs and the achieved advancement can be easily transferred into different modalities.

\section{Conclusion}
In this paper, we introduce a new generative model to scale the DPF model for high-resolution data generation, while inheriting its modality-agnostic flexibility.
Our method involves (1) a new view-wise sampling algorithm based on high-resolution signals;
(2) a conditioning mechanism that leverages view-level noise and text descriptions as inductive bias.
Experimental results demonstrate its effectiveness in various modalities including image, video, and 3D viewpoint.

\setcitestyle{numbers}
\bibliographystyle{plainnat}
\bibliography{neurips_2023/references, neurips_2023/others} 

\begin{thebibliography}{63}
\providecommand{\natexlab}[1]{#1}
\providecommand{\url}[1]{\texttt{#1}}
\expandafter\ifx\csname urlstyle\endcsname\relax
  \providecommand{\doi}[1]{doi: #1}\else
  \providecommand{\doi}{doi: \begingroup \urlstyle{rm}\Url}\fi

\bibitem[Bain et~al.(2021)Bain, Nagrani, Varol, and Zisserman]{bain2021frozen}
Max Bain, Arsha Nagrani, G{\"u}l Varol, and Andrew Zisserman.
\newblock Frozen in time: A joint video and image encoder for end-to-end
  retrieval.
\newblock In \emph{Proceedings of the IEEE/CVF International Conference on
  Computer Vision}, pages 1728--1738, 2021.

\bibitem[Balaji et~al.(2019)Balaji, Min, Bai, Chellappa, and
  Graf]{balaji2019conditional}
Yogesh Balaji, Martin~Renqiang Min, Bing Bai, Rama Chellappa, and Hans~Peter
  Graf.
\newblock Conditional gan with discriminative filter generation for
  text-to-video synthesis.
\newblock In \emph{IJCAI}, volume~1, page~2, 2019.

\bibitem[Bauer et~al.(2023)Bauer, Dupont, Brock, Rosenbaum, Schwarz, and
  Kim]{bauer2023spatial}
Matthias Bauer, Emilien Dupont, Andy Brock, Dan Rosenbaum, Jonathan Schwarz,
  and Hyunjik Kim.
\newblock Spatial functa: Scaling functa to imagenet classification and
  generation.
\newblock \emph{arXiv preprint arXiv:2302.03130}, 2023.

\bibitem[Blattmann et~al.(2023)Blattmann, Rombach, Ling, Dockhorn, Kim, Fidler,
  and Kreis]{blattmann2023align}
Andreas Blattmann, Robin Rombach, Huan Ling, Tim Dockhorn, Seung~Wook Kim,
  Sanja Fidler, and Karsten Kreis.
\newblock Align your latents: High-resolution video synthesis with latent
  diffusion models.
\newblock \emph{arXiv preprint arXiv:2304.08818}, 2023.

\bibitem[Brock et~al.(2018)Brock, Donahue, and Simonyan]{brock2018large}
Andrew Brock, Jeff Donahue, and Karen Simonyan.
\newblock Large scale gan training for high fidelity natural image synthesis.
\newblock \emph{arXiv preprint arXiv:1809.11096}, 2018.

\bibitem[Brown et~al.(2020)Brown, Mann, Ryder, Subbiah, Kaplan, Dhariwal,
  Neelakantan, Shyam, Sastry, Askell, et~al.]{brown2020language}
Tom Brown, Benjamin Mann, Nick Ryder, Melanie Subbiah, Jared~D Kaplan, Prafulla
  Dhariwal, Arvind Neelakantan, Pranav Shyam, Girish Sastry, Amanda Askell,
  et~al.
\newblock Language models are few-shot learners.
\newblock \emph{Advances in neural information processing systems},
  33:\penalty0 1877--1901, 2020.

\bibitem[Chan et~al.(2022)Chan, Lin, Chan, Nagano, Pan, De~Mello, Gallo,
  Guibas, Tremblay, Khamis, et~al.]{chan2022efficient}
Eric~R Chan, Connor~Z Lin, Matthew~A Chan, Koki Nagano, Boxiao Pan, Shalini
  De~Mello, Orazio Gallo, Leonidas~J Guibas, Jonathan Tremblay, Sameh Khamis,
  et~al.
\newblock Efficient geometry-aware 3d generative adversarial networks.
\newblock In \emph{Proceedings of the IEEE/CVF Conference on Computer Vision
  and Pattern Recognition}, pages 16123--16133, 2022.

\bibitem[Chan et~al.(2023)Chan, Nagano, Chan, Bergman, Park, Levy, Aittala,
  De~Mello, Karras, and Wetzstein]{chan2023generative}
Eric~R Chan, Koki Nagano, Matthew~A Chan, Alexander~W Bergman, Jeong~Joon Park,
  Axel Levy, Miika Aittala, Shalini De~Mello, Tero Karras, and Gordon
  Wetzstein.
\newblock Generative novel view synthesis with 3d-aware diffusion models.
\newblock \emph{arXiv preprint arXiv:2304.02602}, 2023.

\bibitem[Chang et~al.(2015)Chang, Funkhouser, Guibas, Hanrahan, Huang, Li,
  Savarese, Savva, Song, Su, et~al.]{chang2015shapenet}
Angel~X Chang, Thomas Funkhouser, Leonidas Guibas, Pat Hanrahan, Qixing Huang,
  Zimo Li, Silvio Savarese, Manolis Savva, Shuran Song, Hao Su, et~al.
\newblock Shapenet: An information-rich 3d model repository.
\newblock \emph{arXiv preprint arXiv:1512.03012}, 2015.

\bibitem[Chen et~al.(2020)Chen, Radford, Child, Wu, Jun, Luan, and
  Sutskever]{chen2020generative}
Mark Chen, Alec Radford, Rewon Child, Jeffrey Wu, Heewoo Jun, David Luan, and
  Ilya Sutskever.
\newblock Generative pretraining from pixels.
\newblock In \emph{International conference on machine learning}, pages
  1691--1703. PMLR, 2020.

\bibitem[Devlin et~al.(2018)Devlin, Chang, Lee, and Toutanova]{devlin2018bert}
Jacob Devlin, Ming-Wei Chang, Kenton Lee, and Kristina Toutanova.
\newblock Bert: Pre-training of deep bidirectional transformers for language
  understanding.
\newblock \emph{arXiv preprint arXiv:1810.04805}, 2018.

\bibitem[Dhariwal and Nichol(2021)]{dhariwal2021diffusion}
Prafulla Dhariwal and Alexander Nichol.
\newblock Diffusion models beat gans on image synthesis.
\newblock \emph{Advances in Neural Information Processing Systems},
  34:\penalty0 8780--8794, 2021.

\bibitem[Du et~al.(2021)Du, Collins, Tenenbaum, and Sitzmann]{du2021learning}
Yilun Du, Katie Collins, Josh Tenenbaum, and Vincent Sitzmann.
\newblock Learning signal-agnostic manifolds of neural fields.
\newblock \emph{Advances in Neural Information Processing Systems},
  34:\penalty0 8320--8331, 2021.

\bibitem[Dupont et~al.(2021)Dupont, Teh, and Doucet]{dupont2021generative}
Emilien Dupont, Yee~Whye Teh, and Arnaud Doucet.
\newblock Generative models as distributions of functions.
\newblock \emph{arXiv preprint arXiv:2102.04776}, 2021.

\bibitem[Dupont et~al.(2022{\natexlab{a}})Dupont, Kim, Eslami, Rezende, and
  Rosenbaum]{dupont_data_2022}
Emilien Dupont, Hyunjik Kim, S.~M.~Ali Eslami, Danilo Rezende, and Dan
  Rosenbaum.
\newblock From data to functa: {Your} data point is a function and you can
  treat it like one, November 2022{\natexlab{a}}.
\newblock URL \url{http://arxiv.org/abs/2201.12204}.
\newblock arXiv:2201.12204 [cs].

\bibitem[Dupont et~al.(2022{\natexlab{b}})Dupont, Kim, Eslami, Rezende, and
  Rosenbaum]{dupont2022data}
Emilien Dupont, Hyunjik Kim, SM~Eslami, Danilo Rezende, and Dan Rosenbaum.
\newblock From data to functa: Your data point is a function and you should
  treat it like one.
\newblock \emph{arXiv preprint arXiv:2201.12204}, 2022{\natexlab{b}}.

\bibitem[Dutordoir et~al.(2022)Dutordoir, Saul, Ghahramani, and
  Simpson]{dutordoir2022neural}
Vincent Dutordoir, Alan Saul, Zoubin Ghahramani, and Fergus Simpson.
\newblock Neural diffusion processes.
\newblock \emph{arXiv preprint arXiv:2206.03992}, 2022.

\bibitem[Esser et~al.(2021)Esser, Rombach, and Ommer]{esser2021taming}
Patrick Esser, Robin Rombach, and Bjorn Ommer.
\newblock Taming transformers for high-resolution image synthesis.
\newblock In \emph{Proceedings of the IEEE/CVF conference on computer vision
  and pattern recognition}, pages 12873--12883, 2021.

\bibitem[Goodfellow et~al.(2020)Goodfellow, Pouget-Abadie, Mirza, Xu,
  Warde-Farley, Ozair, Courville, and Bengio]{goodfellow2020generative}
Ian Goodfellow, Jean Pouget-Abadie, Mehdi Mirza, Bing Xu, David Warde-Farley,
  Sherjil Ozair, Aaron Courville, and Yoshua Bengio.
\newblock Generative adversarial networks.
\newblock \emph{Communications of the ACM}, 63\penalty0 (11):\penalty0
  139--144, 2020.

\bibitem[Gordon et~al.(2020)Gordon, Bruinsma, Foong, Requeima, Dubois, and
  Turner]{Gordon2020Convolutional}
Jonathan Gordon, Wessel~P. Bruinsma, Andrew Y.~K. Foong, James Requeima, Yann
  Dubois, and Richard~E. Turner.
\newblock Convolutional conditional neural processes.
\newblock In \emph{International Conference on Learning Representations}, 2020.
\newblock URL \url{https://openreview.net/forum?id=Skey4eBYPS}.

\bibitem[Ha et~al.(2016)Ha, Dai, and Le]{ha2016hypernetworks}
David Ha, Andrew Dai, and Quoc~V Le.
\newblock Hypernetworks.
\newblock \emph{arXiv preprint arXiv:1609.09106}, 2016.

\bibitem[Han et~al.(2022{\natexlab{a}})Han, Ren, Lee, Barbieri, Olszewski,
  Minaee, Metaxas, and Tulyakov]{han2022show}
Ligong Han, Jian Ren, Hsin-Ying Lee, Francesco Barbieri, Kyle Olszewski,
  Shervin Minaee, Dimitris Metaxas, and Sergey Tulyakov.
\newblock Show me what and tell me how: Video synthesis via multimodal
  conditioning.
\newblock In \emph{Proceedings of the IEEE/CVF Conference on Computer Vision
  and Pattern Recognition}, pages 3615--3625, 2022{\natexlab{a}}.

\bibitem[Han et~al.(2022{\natexlab{b}})Han, Ren, Lee, Barbieri, Olszewski,
  Minaee, Metaxas, and Tulyakov]{han_show_2022}
Ligong Han, Jian Ren, Hsin-Ying Lee, Francesco Barbieri, Kyle Olszewski,
  Shervin Minaee, Dimitris Metaxas, and Sergey Tulyakov.
\newblock Show {Me} {What} and {Tell} {Me} {How}: {Video} {Synthesis} via
  {Multimodal} {Conditioning}, March 2022{\natexlab{b}}.
\newblock URL \url{http://arxiv.org/abs/2203.02573}.
\newblock arXiv:2203.02573 [cs].

\bibitem[He et~al.(2022)He, Yang, Zhang, Shan, and Chen]{he_latent_2022}
Yingqing He, Tianyu Yang, Yong Zhang, Ying Shan, and Qifeng Chen.
\newblock Latent {Video} {Diffusion} {Models} for {High}-{Fidelity} {Video}
  {Generation} with {Arbitrary} {Lengths}.
\newblock \emph{arXiv preprint arXiv:2211.13221}, 2022.

\bibitem[Hersbach et~al.(2019)Hersbach, Bell, Berrisford, Biavati, Hor{\'a}nyi,
  Mu{\~n}oz~Sabater, Nicolas, Peubey, Radu, Rozum, et~al.]{hersbach2019era5}
H~Hersbach, B~Bell, P~Berrisford, G~Biavati, A~Hor{\'a}nyi,
  J~Mu{\~n}oz~Sabater, J~Nicolas, C~Peubey, R~Radu, I~Rozum, et~al.
\newblock Era5 monthly averaged data on single levels from 1979 to present.
\newblock \emph{Copernicus Climate Change Service (C3S) Climate Data Store
  (CDS)}, 10:\penalty0 252--266, 2019.

\bibitem[Heusel et~al.(2017)Heusel, Ramsauer, Unterthiner, Nessler, and
  Hochreiter]{heusel2017gans}
Martin Heusel, Hubert Ramsauer, Thomas Unterthiner, Bernhard Nessler, and Sepp
  Hochreiter.
\newblock Gans trained by a two time-scale update rule converge to a local nash
  equilibrium.
\newblock \emph{Advances in neural information processing systems}, 30, 2017.

\bibitem[Ho et~al.(2020)Ho, Jain, and Abbeel]{ho2020denoising}
Jonathan Ho, Ajay Jain, and Pieter Abbeel.
\newblock Denoising diffusion probabilistic models.
\newblock \emph{Advances in Neural Information Processing Systems},
  33:\penalty0 6840--6851, 2020.

\bibitem[Ho et~al.(2022{\natexlab{a}})Ho, Chan, Saharia, Whang, Gao, Gritsenko,
  Kingma, Poole, Norouzi, and Fleet]{ho_imagen_2022}
Jonathan Ho, William Chan, Chitwan Saharia, Jay Whang, Ruiqi Gao, Alexey
  Gritsenko, Diederik~P. Kingma, Ben Poole, Mohammad Norouzi, and David~J.
  Fleet.
\newblock Imagen video: {High} definition video generation with diffusion
  models.
\newblock \emph{arXiv preprint arXiv:2210.02303}, 2022{\natexlab{a}}.

\bibitem[Ho et~al.(2022{\natexlab{b}})Ho, Salimans, Gritsenko, Chan, Norouzi,
  and Fleet]{ho_video_2022}
Jonathan Ho, Tim Salimans, Alexey Gritsenko, William Chan, Mohammad Norouzi,
  and David~J. Fleet.
\newblock Video diffusion models.
\newblock In \emph{Advances in {Neural} {Information} {Processing} {Systems}},
  2022{\natexlab{b}}.

\bibitem[Hong et~al.(2023)Hong, Ding, Zheng, Liu, and Tang]{hong2022cogvideo}
Wenyi Hong, Ming Ding, Wendi Zheng, Xinghan Liu, and Jie Tang.
\newblock Cogvideo: Large-scale pretraining for text-to-video generation via
  transformers.
\newblock In \emph{International Conference on Learning Representations}, 2023.

\bibitem[Jaegle et~al.(2021)Jaegle, Borgeaud, Alayrac, Doersch, Ionescu, Ding,
  Koppula, Zoran, Brock, Shelhamer, et~al.]{jaegle2021perceiver}
Andrew Jaegle, Sebastian Borgeaud, Jean-Baptiste Alayrac, Carl Doersch, Catalin
  Ionescu, David Ding, Skanda Koppula, Daniel Zoran, Andrew Brock, Evan
  Shelhamer, et~al.
\newblock Perceiver io: A general architecture for structured inputs \&
  outputs.
\newblock \emph{arXiv preprint arXiv:2107.14795}, 2021.

\bibitem[Karras et~al.(2017)Karras, Aila, Laine, and
  Lehtinen]{karras2017progressive}
Tero Karras, Timo Aila, Samuli Laine, and Jaakko Lehtinen.
\newblock Progressive growing of gans for improved quality, stability, and
  variation.
\newblock \emph{arXiv preprint arXiv:1710.10196}, 2017.

\bibitem[Karras et~al.(2019)Karras, Laine, and Aila]{karras2019style}
Tero Karras, Samuli Laine, and Timo Aila.
\newblock A style-based generator architecture for generative adversarial
  networks.
\newblock In \emph{Proceedings of the IEEE/CVF conference on computer vision
  and pattern recognition}, pages 4401--4410, 2019.

\bibitem[Kingma and Welling(2013)]{kingma2013auto}
Diederik~P Kingma and Max Welling.
\newblock Auto-encoding variational bayes.
\newblock \emph{arXiv preprint arXiv:1312.6114}, 2013.

\bibitem[Kulh{\'a}nek et~al.(2022)Kulh{\'a}nek, Derner, Sattler, and
  Babu{\v{s}}ka]{kulhanek2022viewformer}
Jon{\'a}{\v{s}} Kulh{\'a}nek, Erik Derner, Torsten Sattler, and Robert
  Babu{\v{s}}ka.
\newblock Viewformer: Nerf-free neural rendering from few images using
  transformers.
\newblock In \emph{Computer Vision--ECCV 2022: 17th European Conference, Tel
  Aviv, Israel, October 23--27, 2022, Proceedings, Part XV}, pages 198--216.
  Springer, 2022.

\bibitem[Kynk{\"a}{\"a}nniemi et~al.(2019)Kynk{\"a}{\"a}nniemi, Karras, Laine,
  Lehtinen, and Aila]{kynkaanniemi2019improved}
Tuomas Kynk{\"a}{\"a}nniemi, Tero Karras, Samuli Laine, Jaakko Lehtinen, and
  Timo Aila.
\newblock Improved precision and recall metric for assessing generative models.
\newblock \emph{Advances in Neural Information Processing Systems}, 32, 2019.

\bibitem[Liu et~al.(2015)Liu, Luo, Wang, and Tang]{liu2015deep}
Ziwei Liu, Ping Luo, Xiaogang Wang, and Xiaoou Tang.
\newblock Deep learning face attributes in the wild.
\newblock In \emph{Proceedings of the IEEE international conference on computer
  vision}, pages 3730--3738, 2015.

\bibitem[Mao et~al.(2017)Mao, Li, Xie, Lau, Wang, and
  Paul~Smolley]{mao2017least}
Xudong Mao, Qing Li, Haoran Xie, Raymond~YK Lau, Zhen Wang, and Stephen
  Paul~Smolley.
\newblock Least squares generative adversarial networks.
\newblock In \emph{Proceedings of the IEEE international conference on computer
  vision}, pages 2794--2802, 2017.

\bibitem[Peebles and Xie(2022{\natexlab{a}})]{peebles2022scalable}
William Peebles and Saining Xie.
\newblock Scalable diffusion models with transformers.
\newblock \emph{arXiv preprint arXiv:2212.09748}, 2022{\natexlab{a}}.

\bibitem[Peebles and Xie(2022{\natexlab{b}})]{peebles_scalable_2022}
William Peebles and Saining Xie.
\newblock Scalable {Diffusion} {Models} with {Transformers}.
\newblock \emph{arXiv preprint arXiv:2212.09748}, 2022{\natexlab{b}}.

\bibitem[Quinonero-Candela and Rasmussen(2005)]{quinonero2005unifying}
Joaquin Quinonero-Candela and Carl~Edward Rasmussen.
\newblock A unifying view of sparse approximate gaussian process regression.
\newblock \emph{The Journal of Machine Learning Research}, 6:\penalty0
  1939--1959, 2005.

\bibitem[Radford et~al.(2021)Radford, Kim, Hallacy, Ramesh, Goh, Agarwal,
  Sastry, Askell, Mishkin, Clark, et~al.]{radford2021learning}
Alec Radford, Jong~Wook Kim, Chris Hallacy, Aditya Ramesh, Gabriel Goh,
  Sandhini Agarwal, Girish Sastry, Amanda Askell, Pamela Mishkin, Jack Clark,
  et~al.
\newblock Learning transferable visual models from natural language
  supervision.
\newblock In \emph{International conference on machine learning}, pages
  8748--8763. PMLR, 2021.

\bibitem[Raffel et~al.(2020)Raffel, Shazeer, Roberts, Lee, Narang, Matena,
  Zhou, Li, and Liu]{raffel2020exploring}
Colin Raffel, Noam Shazeer, Adam Roberts, Katherine Lee, Sharan Narang, Michael
  Matena, Yanqi Zhou, Wei Li, and Peter~J Liu.
\newblock Exploring the limits of transfer learning with a unified text-to-text
  transformer.
\newblock \emph{The Journal of Machine Learning Research}, 21\penalty0
  (1):\penalty0 5485--5551, 2020.

\bibitem[Ramesh et~al.(2022)Ramesh, Dhariwal, Nichol, Chu, and
  Chen]{ramesh2022hierarchical}
Aditya Ramesh, Prafulla Dhariwal, Alex Nichol, Casey Chu, and Mark Chen.
\newblock Hierarchical text-conditional image generation with clip latents.
\newblock \emph{arXiv preprint arXiv:2204.06125}, 2022.

\bibitem[Rombach et~al.(2022)Rombach, Blattmann, Lorenz, Esser, and
  Ommer]{rombach_high-resolution_2022}
Robin Rombach, Andreas Blattmann, Dominik Lorenz, Patrick Esser, and Björn
  Ommer.
\newblock High-resolution image synthesis with latent diffusion models.
\newblock In \emph{Proceedings of the {IEEE}/{CVF} {Conference} on {Computer}
  {Vision} and {Pattern} {Recognition}}, pages 10684--10695, 2022.

\bibitem[Rostamzadeh et~al.(2021)Rostamzadeh, Denton, and
  Petrini]{rostamzadeh2021ethics}
Negar Rostamzadeh, Emily Denton, and Linda Petrini.
\newblock Ethics and creativity in computer vision.
\newblock \emph{arXiv preprint arXiv:2112.03111}, 2021.

\bibitem[Sajjadi et~al.(2018)Sajjadi, Bachem, Lucic, Bousquet, and
  Gelly]{sajjadi2018assessing}
Mehdi~SM Sajjadi, Olivier Bachem, Mario Lucic, Olivier Bousquet, and Sylvain
  Gelly.
\newblock Assessing generative models via precision and recall.
\newblock \emph{Advances in neural information processing systems}, 31, 2018.

\bibitem[Singer et~al.(2022)Singer, Polyak, Hayes, Yin, An, Zhang, Hu, Yang,
  Ashual, Gafni, et~al.]{singer2022make}
Uriel Singer, Adam Polyak, Thomas Hayes, Xi~Yin, Jie An, Songyang Zhang, Qiyuan
  Hu, Harry Yang, Oron Ashual, Oran Gafni, et~al.
\newblock Make-a-video: Text-to-video generation without text-video data.
\newblock \emph{arXiv preprint arXiv:2209.14792}, 2022.

\bibitem[Sitzmann et~al.(2019)Sitzmann, Zollh{\"o}fer, and
  Wetzstein]{sitzmann2019scene}
Vincent Sitzmann, Michael Zollh{\"o}fer, and Gordon Wetzstein.
\newblock Scene representation networks: Continuous 3d-structure-aware neural
  scene representations.
\newblock \emph{Advances in Neural Information Processing Systems}, 32, 2019.

\bibitem[Sitzmann et~al.(2020)Sitzmann, Martel, Bergman, Lindell, and
  Wetzstein]{sitzmann2020implicit}
Vincent Sitzmann, Julien Martel, Alexander Bergman, David Lindell, and Gordon
  Wetzstein.
\newblock Implicit neural representations with periodic activation functions.
\newblock \emph{Advances in Neural Information Processing Systems},
  33:\penalty0 7462--7473, 2020.

\bibitem[Song and Ermon(2020)]{song2020improved}
Yang Song and Stefano Ermon.
\newblock Improved techniques for training score-based generative models.
\newblock \emph{Advances in neural information processing systems},
  33:\penalty0 12438--12448, 2020.

\bibitem[Song et~al.(2020)Song, Sohl-Dickstein, Kingma, Kumar, Ermon, and
  Poole]{song2020score}
Yang Song, Jascha Sohl-Dickstein, Diederik~P Kingma, Abhishek Kumar, Stefano
  Ermon, and Ben Poole.
\newblock Score-based generative modeling through stochastic differential
  equations.
\newblock \emph{arXiv preprint arXiv:2011.13456}, 2020.

\bibitem[Tancik et~al.(2020)Tancik, Srinivasan, Mildenhall, Fridovich-Keil,
  Raghavan, Singhal, Ramamoorthi, Barron, and Ng]{tancik2020fourier}
Matthew Tancik, Pratul Srinivasan, Ben Mildenhall, Sara Fridovich-Keil, Nithin
  Raghavan, Utkarsh Singhal, Ravi Ramamoorthi, Jonathan Barron, and Ren Ng.
\newblock Fourier features let networks learn high frequency functions in low
  dimensional domains.
\newblock \emph{Advances in Neural Information Processing Systems},
  33:\penalty0 7537--7547, 2020.

\bibitem[Unterthiner et~al.(2018)Unterthiner, Van~Steenkiste, Kurach, Marinier,
  Michalski, and Gelly]{unterthiner2018towards}
Thomas Unterthiner, Sjoerd Van~Steenkiste, Karol Kurach, Raphael Marinier,
  Marcin Michalski, and Sylvain Gelly.
\newblock Towards accurate generative models of video: A new metric \&
  challenges.
\newblock \emph{arXiv preprint arXiv:1812.01717}, 2018.

\bibitem[Vahdat and Kautz(2020)]{vahdat2020nvae}
Arash Vahdat and Jan Kautz.
\newblock Nvae: A deep hierarchical variational autoencoder.
\newblock \emph{Advances in neural information processing systems},
  33:\penalty0 19667--19679, 2020.

\bibitem[Wu et~al.(2021)Wu, Huang, Zhang, Li, Ji, Yang, Sapiro, and
  Duan]{wu2021godiva}
Chenfei Wu, Lun Huang, Qianxi Zhang, Binyang Li, Lei Ji, Fan Yang, Guillermo
  Sapiro, and Nan Duan.
\newblock Godiva: Generating open-domain videos from natural descriptions.
\newblock \emph{arXiv preprint arXiv:2104.14806}, 2021.

\bibitem[Wu et~al.(2022)Wu, Liang, Ji, Yang, Fang, Jiang, and Duan]{wu2022nuwa}
Chenfei Wu, Jian Liang, Lei Ji, Fan Yang, Yuejian Fang, Daxin Jiang, and Nan
  Duan.
\newblock N{\"u}wa: Visual synthesis pre-training for neural visual world
  creation.
\newblock In \emph{Computer Vision--ECCV 2022: 17th European Conference, Tel
  Aviv, Israel, October 23--27, 2022, Proceedings, Part XVI}, pages 720--736.
  Springer, 2022.

\bibitem[Yu et~al.(2021)Yu, Ye, Tancik, and Kanazawa]{yu2021pixelnerf}
Alex Yu, Vickie Ye, Matthew Tancik, and Angjoo Kanazawa.
\newblock pixelnerf: Neural radiance fields from one or few images.
\newblock In \emph{Proceedings of the IEEE/CVF Conference on Computer Vision
  and Pattern Recognition}, pages 4578--4587, 2021.

\bibitem[Yu et~al.(2023{\natexlab{a}})Yu, Zhu, Jiang, Loy, Cai, and
  Wu]{yu2023celebv}
Jianhui Yu, Hao Zhu, Liming Jiang, Chen~Change Loy, Weidong Cai, and Wayne Wu.
\newblock Celebv-text: A large-scale facial text-video dataset.
\newblock \emph{arXiv preprint arXiv:2303.14717}, 2023{\natexlab{a}}.

\bibitem[Yu et~al.(2023{\natexlab{b}})Yu, Zhu, Jiang, Loy, Cai, and
  Wu]{yu_celebv-text_2023}
Jianhui Yu, Hao Zhu, Liming Jiang, Chen~Change Loy, Weidong Cai, and Wayne Wu.
\newblock {CelebV}-{Text}: {A} {Large}-{Scale} {Facial} {Text}-{Video}
  {Dataset}.
\newblock \emph{arXiv preprint arXiv:2303.14717}, 2023{\natexlab{b}}.

\bibitem[Yu et~al.(2023{\natexlab{c}})Yu, Sohn, Kim, and Shin]{yu2023video}
Sihyun Yu, Kihyuk Sohn, Subin Kim, and Jinwoo Shin.
\newblock Video probabilistic diffusion models in projected latent space.
\newblock \emph{arXiv preprint arXiv:2302.07685}, 2023{\natexlab{c}}.

\bibitem[Zhou et~al.(2022)Zhou, Wang, Yan, Lv, Zhu, and
  Feng]{zhou2022magicvideo}
Daquan Zhou, Weimin Wang, Hanshu Yan, Weiwei Lv, Yizhe Zhu, and Jiashi Feng.
\newblock Magicvideo: Efficient video generation with latent diffusion models.
\newblock \emph{arXiv preprint arXiv:2211.11018}, 2022.

\bibitem[Zhuang et~al.(2023)Zhuang, Abnar, Gu, Schwing, Susskind, and
  Bautista]{zhuang_diffusion_2023}
Peiye Zhuang, Samira Abnar, Jiatao Gu, Alex Schwing, Joshua~M. Susskind, and
  Miguel~Ángel Bautista.
\newblock Diffusion {Probabilistic} {Fields}.
\newblock In \emph{International {Conference} on {Learning} {Representations}},
  2023.

\end{thebibliography}

\newpage
\appendix
\section{Additional Results}
The additional results are located at \url{https://t1-diffusion-model.github.io}.

\section{Ethical Statement}
In this paper, we present a new generative model unifying varies visual content modalities including images, videos, and 3D scenes. While we are excited about the potential applications of our model, we are also acutely aware of the possible risks and challenges associated with its deployment.
Our model's ability to generate realistic videos and 3D scenes could potentially be misused for creating disingenuous data, \emph{a.k,a}, ``DeepFakes''.
We encourage the research community and practitioners to follow privacy-preserving practices when utilizing our model.
We also encourage readers to refer to the Rostamzadeh et al.~\cite{rostamzadeh2021ethics} for an in-depth review of ethics in generating visual contents.

\section{Additional Settings}
\paragraph{Model Details.}
\begin{itemize}
    \item In the interest of maintaining simplicity, we adhere to the methodology outlined by Dhariwal et al.~\cite{dhariwal2021diffusion} and utilize a 256-dimensional frequency embedding to encapsulate input denoising timesteps. This embedding is then refined through a two-layer Multilayer Perceptron (MLP) with Swish (SiLU) activation functions.
    \item Our model aligns with the size configuration of DiT-XL~\cite{peebles_scalable_2022}, which includes retaining the number of transformer blocks (\emph{i.e.} 28), the hidden dimension size of each transformer block (\emph{i.e.}, 1152), and the number of attention heads (\emph{i.e.}, 16).
    \item Our model derives text embeddings employing T5-XXL~\cite{raffel2020exploring}, culminating in a fixed length token sequence (\emph{i.e.}, 256) which matches the length of the noisy tokens. To further process each text embedding token, our model compresses them via a single layer MLP, which has a hidden dimension size identical to that of the transformer block.
\end{itemize}

\paragraph{Diffusion Process Details.} Our model uses classifier-free guidance in the backward process with a fixed scale of 8.5. To keep consistency with DiT-XL~\cite{peebles2022scalable}, we only applied guidance to the first three channels of each denoised token.

\paragraph{3D Geometry Rendering Settings.}
Following the settings of pixelNeRF~\cite{yu2021pixelnerf}, we render each car voxel into 128 random views for training models and testing.
However, the original setting puts camera far away from the objects and hence results in two many blank areas in the rendered views.
We empirically find that these blank areas hurts the diffusion model performance since the noise becomes obvious in blank area and can be easily inferred by diffusion models, which degrades the distribution modeling capability of diffusion models.

\begin{figure}[htbp]
    \centering
    \begin{subfigure}{.49\textwidth}
    \includegraphics[width=\textwidth]{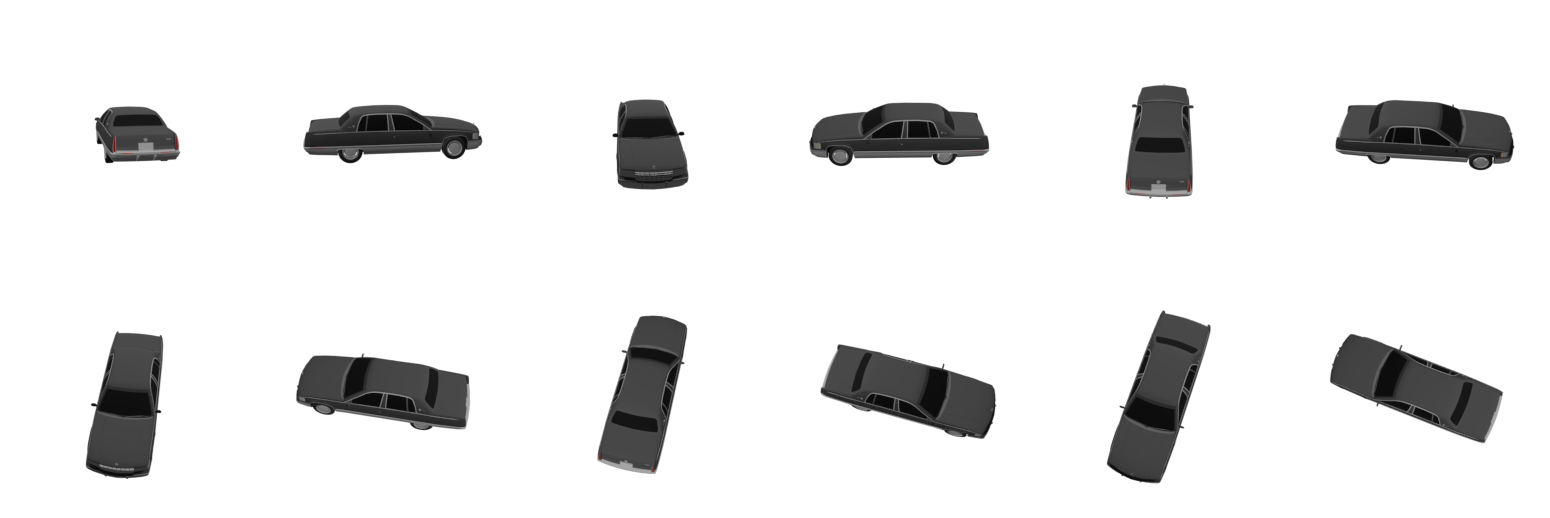}
    \caption{pixelNeRF~\cite{yu2021pixelnerf} rendering}
    \end{subfigure}
    \begin{subfigure}{.49\textwidth}
    \includegraphics[width=\textwidth]{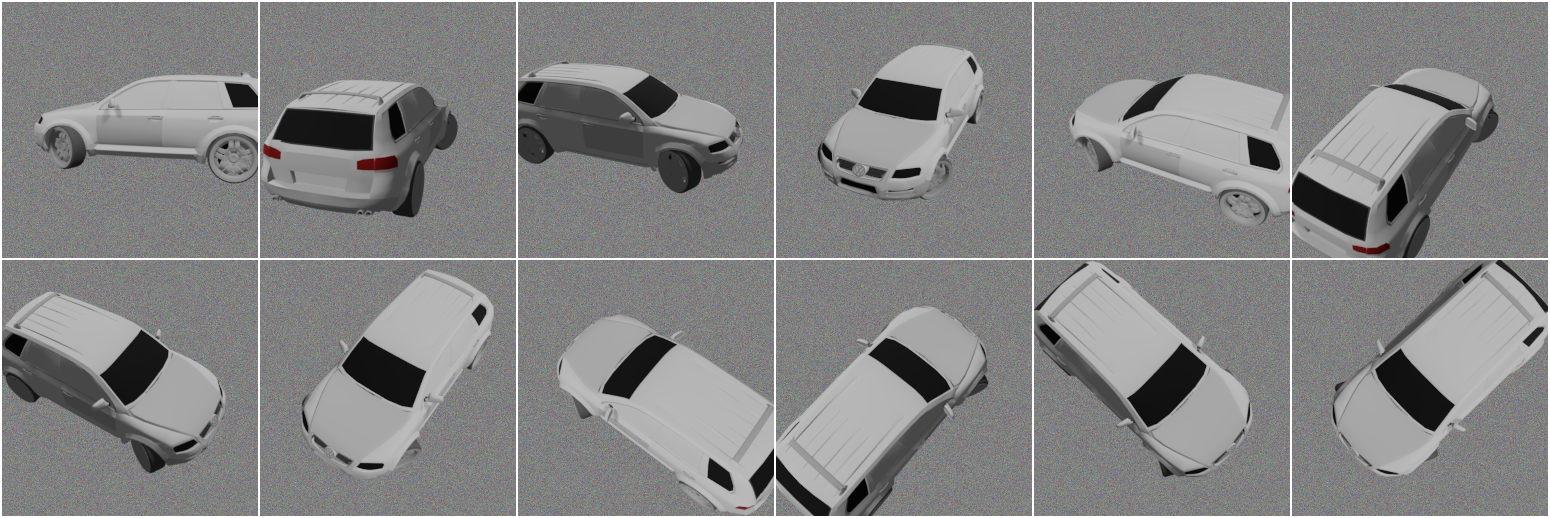}
    \caption{ours rendering}
    \end{subfigure}
    \caption{Visual comparisons of different 3D geometry rendering settings.}
    \label{fig:abrendering}
\end{figure}

To overcome this, we first randomly fill the blank area with Gaussian noise $\mathcal{N}(0, 0.1)$ without overlapping the 3D geometry.
We then move the camera in the z-axis from 4.0 into 3.0, which is closer to the object than the previous one.
During testing, we use the same settings as pixelNeRF and remove the noise according to the mask.
For straightforward understand their difference, we visualized their rendered results in Fig.~\ref{fig:abrendering}.

\begin{table}[h]
\begin{center}
\begin{tabular}{r c c c}
\toprule
\bf{Hyper-parameter} & \bf{CelebA-HQ~\cite{liu2015deep}} & \bf{CelebV-Text~\cite{yu_celebv-text_2023}} & \bf{ShapeNet~\cite{chang2015shapenet}} \\
\midrule
train res. & 64$\times$64 & 256$\times$256$\times$128 & \thead{128$\times$128$\times$128 \\ 256$\times$256$\times$128 (upsampled)}  \\
eval res. & 64$\times$64 & 256$\times$256$\times$128 & \thead{128$\times$128$\times$251 \\ 256$\times$256$\times$251} \\
\# dim coordinates & $2$ & $3$ & $6$ \\
\# dim signal & $3$  & $3$ &  $3$\\
\midrule
\# freq pos. embed & $10$ & $10$ & $10$ \\
\# freq pos. embed $t$ & $64$ & $64$ & $64$ \\
\midrule
\#blocks & $28$ &  $28$ & $28$ \\
\#block latents & $1152$ & $1152$ & $1152$ \\
\#self attention heads & $16$ & $16$ & $16$ \\
\midrule
batch size & $128$ & $128$ & $128$ \\
lr  & $1e-4$ & $1e-4$ & $1e-4$ \\
epochs  & $400$ & $400$ & $1200$ \\
\bottomrule
\end{tabular}
\end{center}
\caption{Hyperparameters and settings on different datasets.}
\label{tab:hyper}
\end{table}

\section{Additional Dataset Details}

In the subsequent sections, we present the datasets utilized for conducting our experiments.
We empirically change the size settings of our model as shown in Tab~\ref{tab:hyper}.

\begin{itemize}
    \item \textbf{CelebV-Text~\cite{yu_celebv-text_2023}.} Due to the unavailability of some videos in the released dataset, we utilize the first 60,000 downloadable videos for training our model. For videos that contain more than 128 frames, we uniformly select 128 frames. Conversely, for videos with fewer than 128 frames, we move to the next video, following the order of their names, until we identify a video that meets the required length of 128 frames.
    \item \textbf{ShapeNet~\cite{chang2015shapenet}.} The conventional methods in DPF~\cite{zhuang_diffusion_2023} and GEM~\cite{du2021learning} generally involve training models on the ShapeNet dataset, wherein each object is depicted as a voxel grid at a resolution of $64^3$. However, our model distinguishes itself by relying on view-level pairs, thereby adopting strategies utilized by innovative view synthesis methods like pixelNeRF~\cite{yu2021pixelnerf} and GeNVS~\cite{chan2023generative}. To specify, we conduct training on the car classes of ShapeNet, which encompasses 2,458 cars, each demonstrated with 128 renderings randomly scattered across the surface of a sphere.
    
    Moreover, it's worth noting that our model refrains from directly leveraging the text descriptions of the car images. Instead, it conditions on the CLIP embedding~\cite{radford2021learning} of car images for linguistic guidance. This approach circumvents the potential accumulation of errors that might occur during the text-to-image transformation process.
\end{itemize}

\section{Additional Experimental Details}
\paragraph{Video Generation Metrics Settings.}
In video generation, we use FVD~\cite{unterthiner2018towards}\footnote{FVD is implemented in \url{https://github.com/sihyun-yu/DIGAN}} to evaluate the video spatial-temporal coherency, FID~\cite{heusel2017gans}\footnote{FID is implemented in \url{https://github.com/toshas/torch-fidelity}} to evaluate the frame quality, and CLIPSIM~\cite{radford2021learning}\footnote{CLIPSIM is implemented in \url{https://github.com/Lightning-AI/torchmetrics}} to evaluate relevance between the generated video and input text.
As all metrics are sensitive to data scale during testing, we randomly select 2,048 videos from the test data and generate results as the ``real'' and ``fake'' part in our metric experiments.
For FID, we uniformly sample 4 frames from each video and use a total of 8,192 images.
For CLIPSIM, we calculate the average score across all frames.
We use the ``openai/clip-vit-large-patch14'' model for extracting features in CLIPSIM calculation.

\end{document}